\documentclass[lettersize,journal]{IEEEtran}
\usepackage{amsmath,amsfonts}
\usepackage{array}
\usepackage[caption=false,font=normalsize,labelfont=sf,textfont=sf]{subfig}
\usepackage{textcomp}
\usepackage{stfloats}
\usepackage{url}
\usepackage{verbatim}
\usepackage{graphicx}
\usepackage{cite}
\begin{document}

\title{WGAST: Weakly-Supervised Generative Network for Daily 10 m Land Surface Temperature Estimation via  Spatio-Temporal Fusion}

\author{Sofiane~Bouaziz,~\IEEEmembership{Student Member,~IEEE,}
Adel~Hafiane,~\IEEEmembership{Member,~IEEE,}
Raphaël~Canals,
Rachid~Nedjai
        
\thanks{This work was supported by Orléans Métropole and Région Centre-Val de Loire. (Corresponding author: Sofiane Bouaziz).}
\thanks{Sofiane Bouaziz is with INSA CVL, PRISME UR 4229, Bourges, 18022, Centre Val de Loire, France (e-mail: sofiane.bouaziz@insa-cvl.fr).}%
\thanks{Sofiane Bouaziz, Raphaël Canals, and Adel Hafiane are with Université d’Orléans, PRISME UR 4229, Orléans, 45067, Centre Val de Loire, France (e-mail: raphael.canals@univ-orleans.fr; adel.hafiane@univ-orleans.fr).}%
\thanks{Rachid Nedjai is with Université d'Orléans, CEDETE, UR 1210, Orléans, 45067, Centre Val de Loire, France (e-mail: rachid.nedjai@univ-orleans.fr).}
}



\maketitle

\begin{abstract}
Urbanization, climate change, and agricultural stress are increasing the demand for precise and timely environmental monitoring. Land Surface Temperature (LST) is a key variable in this context and is retrieved from remote sensing satellites. However, these systems face a trade-off between spatial and temporal resolution. While spatio-temporal fusion methods offer promising solutions, few have addressed the estimation of daily LST at 10 m resolution. In this study, we present WGAST, a weakly-supervised generative network for daily 10 m LST estimation via spatio-temporal fusion of Terra MODIS, Landsat 8, and Sentinel-2. WGAST is the first end-to-end deep learning framework designed for this task. It adopts a conditional generative adversarial architecture, with a generator composed of four stages: feature extraction, fusion, LST reconstruction, and noise suppression. The first stage employs a set of encoders to extract multi-level latent representations from the inputs, which are then fused in the second stage using cosine similarity, normalization, and temporal attention mechanisms. The third stage decodes the fused features into high-resolution LST, followed by a Gaussian filter to suppress high-frequency noise. Training follows a weakly supervised strategy based on physical averaging principles and reinforced by a PatchGAN discriminator. Experiments demonstrate that WGAST outperforms existing methods in both quantitative and qualitative evaluations. Compared to the best-performing baseline, on average, WGAST reduces RMSE by 17.05\% and improves SSIM by 4.22\%. Furthermore, WGAST effectively captures fine-scale thermal patterns, as validated against near-surface air temperature measurements from 33 near-ground sensors. The code is available at \url{https://github.com/Sofianebouaziz1/WGAST.git}.
\end{abstract}

\begin{IEEEkeywords}
Spatio-temporal fusion, land surface temperature, spatial resolution, temporal resolution, generative adversarial networks.
\end{IEEEkeywords}

\section{Introduction}
\IEEEPARstart{C}{limate} change, rapid urbanization, and increasing environmental pressures are reshaping the dynamics of our planet with unprecedented speed~\cite{folke2021our}. Cities are expanding both in size and density, which makes issues like the Urban Heat Island (UHI) effect even worse~\cite{singh2017impact}. At the same time, unpredictable weather patterns and changing ecosystems demand consistent attention and monitoring~\cite{sparrow2020effective}. These phenomena are spatially and temporally complex, and tackling them requires reliable, fine-grained geospatial information with dense temporal frequency~\cite{chu2024fine}. Such information is essential for sustainable urban planning~\cite{li2023machine}, public health~\cite{estrela2023remote}, disaster response~\cite{kerle2024disasters}, and climate adaptation~\cite{sirmacek2022remote}. In this context, access to high-resolution environmental data is not only beneficial but crucial for making timely and informed decisions~\cite{bouaziz2024deep}.

\begin{figure}[h]
  \centering
  \includegraphics[width=0.4\textwidth]
  {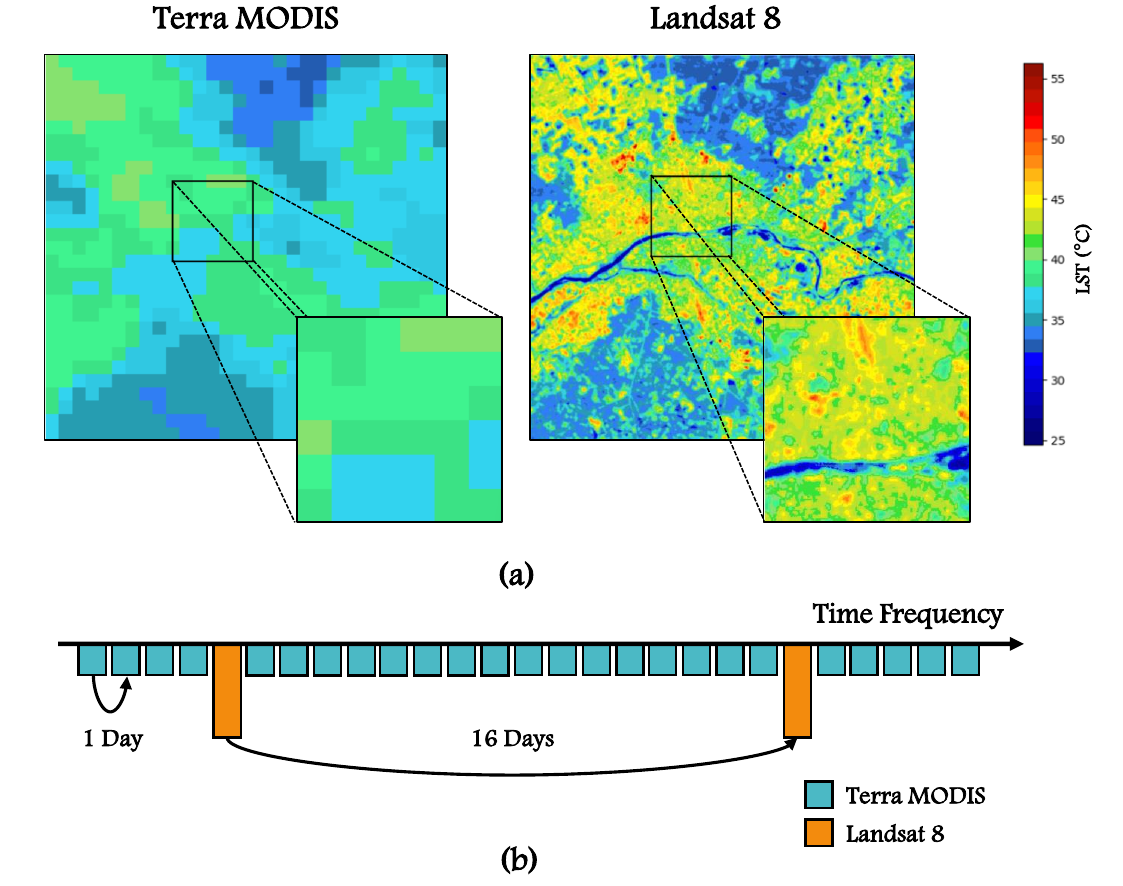}
    \caption{LST resolution comparison over Orléans Métropole, France. (a) Spatial resolution difference between Terra MODIS and Landsat 8 on 13 Aug 2022. (b) Temporal resolution illustrated by the satellites’ revisit frequencies.}
    
  \label{fig:spatial_resolution}
\end{figure}

Land Surface Temperature (LST) is a key geophysical variable for understanding and analyzing diverse environmental challenges. It represents the radiative skin temperature of the land surface, reflecting interactions among solar radiation, land cover, and atmospheric conditions~\cite{HULLEY201957}. LST is widely used in climate change studies~\cite{hall2012satellite}, natural resource management~\cite{luyssaert2014land}, and urban planning~\cite{maimaitiyiming2014effects}. It is recognized by NASA~\cite{king1999eos} and the Global Climate Observing System~\cite{hollmann2013esa} as an essential Earth system and climate variable. Remote sensing (RS) satellites are the primary means for monitoring LST at regional to global scales~\cite{li2013satellite}. They retrieve LST via thermal infrared (TIR) sensors that measure emitted radiation from the Earth's surface~\cite{ZHANG2023129}. Current RS satellites face a trade-off between spatial and temporal resolution~\cite{ZHANG2023129,20201Shunlin}.
Spatial resolution refers to the level of surface detail captured within each LST pixel~\cite{ZHANG2023129}, while temporal resolution indicates how often LST data are acquired for the same geographic area over time~\cite{20201Shunlin}. For example, Terra MODIS provides daily LST at 1 km resolution, whereas Landsat 8 offers 30 m resolution but revisits every 16 days the same area. Fig.\ref{fig:spatial_resolution} illustrates this trade-off, by showing that fine spatial features, such as the Loire River, France’s longest, are visible on Landsat 8 but not on Terra MODIS, and that 16 daily Terra MODIS acquisitions occur between two Landsat 8 observations. Achieving both high spatial and temporal resolution remains challenging but essential for capturing dynamic phenomena.

To overcome this limitation, Spatio-Temporal Fusion (STF) techniques have emerged as a promising solution. They generate high spatial resolution satellite observations at finer temporal frequencies by combining data from RS satellites with complementary characteristics~\cite{zhu2018spatiotemporal}, typically one with high spatial but low temporal resolution (HSLT) and another with low spatial but high temporal resolution (LSHT)~\cite{song2018spatiotemporal}. STF was initially developed for fine-resolution surface reflectance (SR) imagery~\cite{zhu2018spatiotemporal, swain2024spatio}, but has proven effective for LST estimation~\cite{bouaziz2024deep}. Existing STF methods fall into two main categories: traditional and learning-based. Traditional methods include weighted-based, unmixing-based, and hybrid approaches, with weighted-based being the most common.  They predict fine-resolution images using spectrally similar neighboring pixels. Notable examples include STARFM~\cite{gao2006blending}, which uses moving windows and spatial similarity to estimate high-resolution observations. Its enhanced version, ESTARFM~\cite{zhu2010enhanced}, further improves fusion by distinguishing between mixed and pure pixels. Although designed for SR data, these methods have been adapted for LST.  For instance, Liu et al.~\cite{liu2012enhancing} applied STARFM to fuse Terra MODIS and ASTER data and generate ASTER-like LST, and Ma et al.~\cite{ma2018estimation} employed ESTARFM to blend Terra MODIS and Landsat for producing 100 m LST. However, traditional methods rely on the assumption that surface changes are linear over time. This assumption may hold for SR, but fails to capture the non-linear and dynamic nature of LST~\cite{mohamadi2019normalized}. Consequently, traditional methods often struggle to model the spatio-temporal variability of LST with sufficient accuracy. To address this, Yu et al.~\cite{yu2023generating} proposed unbiased ESTARFM, an improved variant of ESTARFM to generate daily 100 m LST while accounting for high temporal dynamics. It introduces a bias correction to reduce temperature deviations between coarse and fine observations. Gao et al. ~\cite{gao2024cpmf} also applied ESTARFM to produce daily 30 m LST and refined it using a random forest with auxiliary variables, such as spectral indices and topography.

Learning-based methods remove the linear assumption by using data-driven models to capture complex spatio-temporal relationships between LSHT and HSLT LST. With advances in deep learning (DL), these approaches have become the most widely used for STF. For example, Yin et al.~\cite{yin2020spatiotemporal} proposed STTFN, a residual multiscale convolutional neural network to estimate fine-resolution LST. Chen et al.~\cite{chen2022spatiotemporal} employed a conditional variational autoencoder framework to generate fine-resolution LST. More recently, Hu et al.~\cite{hu2025two} developed a two-stage hierarchical fusion model based on the Swin Transformer for enhanced LST estimation.

Among recent trends, generative adversarial networks (GANs)~\cite{goodfellow2014generative} have shown notable success in image generation, including STF. In this setting, the generator produces high-resolution fused images, while the discriminator distinguishes them from real high-resolution samples, thus improving realism and fidelity. Several GAN-based STF approaches have been proposed. For example, STFGAN~\cite{zhang2020remote} frames the STF task as a super-resolution problem. GAN-STFM~\cite{tan2021flexible} extends this approach by introducing conditional constraints, which enable a more flexible and context-aware fusion process. CycleGAN-STF~\cite{chen2020cyclegan} treats STF as a data augmentation task and selects the most informative image from cycle-consistent outputs as the final fusion result. MLFF-GAN~\cite{song2022mlff} adopts a multilevel conditional GAN (cGAN) to model spatio-temporal dependencies. However, these methods were designed and validated for SR datasets. Direct application to LST is challenging due to fundamentally different spatio-temporal dynamics~\cite{wu2021spatially}. SR changes with vegetation and seasons, but its daily variation is limited by slow solar illumination. In contrast, LST can vary rapidly, even hourly, due to weather, solar radiation, wind, and other environmental factors~\cite{prata1995thermal}. Spatially, LST can vary sharply over short distances (e.g., streets, roofs, vegetation), whereas SR changes more gradually with vegetation and seasonal cycles~\cite{prata1995thermal}. Consequently, GAN-based STF methods require architectural and methodological adjustments to handle LST’s rapid spatio-temporal variability.

All of the aforementioned STF methods generate fused LST at the minimum spatial resolution of the input satellites, which, at best, is 30 m from Landsat 8~\cite{roy2014landsat}. Thus, both traditional and learning-based STF methods must rely on Landsat 8 as the highest-resolution thermal reference, which restricts their fusion outputs to its native spatial scale. Moreover, most DL-based STF architectures are designed for specific sensor pairs (e.g., Terra MODIS–Landsat 8) and do not generalize easily to heterogeneous optical sensors. While sufficient for some applications, the 30 m resolution is inadequate for studies requiring very high spatial precision, such as UHI analysis, which requires high spatial resolution around 10 m to capture intraurban temperature variations and narrow features like streets and small green spaces~\cite{shi2021urban, zhou2018satellite}. Recent attempts to generate daily 10 m LST include~\cite{mhawej2022daily}, which combines Terra MODIS and Sentinel-2 with robust least squares regression, and~\cite{xiang2025comparative}, which introduces mDTSG to fuse Terra MODIS and Sentinel-2 via a convolution-based moving window. However, both approaches remain traditional STF methods and rely on the same linearity assumptions discussed earlier, which limit their ability to model complex spatio-temporal LST dynamics. Additionally, transitioning from 1 km Terra MODIS to 10 m Sentinel-2 represents a 100$\times$ increase in detail. Direct fusion without an intermediate resolution step amplifies noise, inconsistencies, and may produce unrealistic spatial patterns. Despite this, learning-based DL methods for daily 10 m LST remain largely unexplored. The only attempt, FuseTen~\cite{bouaziz2025fuseten}, uses a hybrid model that combines a cGAN with linear regression by embedding a linear model within the generator, which can introduce artifacts and noise in the generated LST.

In this work, we address the identified research gap by introducing WGAST, a weakly-supervised generative network for daily 10 m LST estimation via STF. To the best of our knowledge, WGAST is the first non-linear end-to-end DL model that simultaneously fuses Terra MODIS, Landsat 8, and Sentinel-2 to produce daily 10 m LST. Our key contributions are summarized as follows:

\begin{itemize}
    \item We propose a feature fusion mechanism that integrates cosine similarity, normalization, and temporal attention to effectively fuse multi-source satellite features.
    \item We introduce a hierarchical approach that employs Landsat 8 as an intermediate resolution bridge to overcome the extreme 1 km to 10 m resolution gap.
    \item We develop a weak supervision strategy that leverages 30 m Landsat 8-derived LST to compensate for the lack of 10 m ground truth.
\end{itemize}

\section{Problem Formulation}
\label{sec:problem_formulation}
The STF problem for estimating daily 10 m LST can be formulated as follows. Let \( X_1 \), \( X_2 \), and \( X_3 \) denote three RS satellites with complementary spatial and temporal resolutions. We define $r_s(\cdot)$ and $r_t(\cdot)$ as the spatial and temporal resolution functions, respectively, and \( \mathrm{TIR}(\cdot) \) as a binary indicator that equals $1$ if the RS satellite provides TIR observations and $0$ otherwise. The following conditions in Equation \ref{eq:X} must hold.
\begin{equation}
\begin{array}{c}
r_s(X_1) > r_s(X_2) > r_s(X_3), \quad r_t(X_1) < r_t(X_2) \\[3pt]
r_s(X_3) = 10\,\text{m}, \quad r_t(X_1) = 1\,\text{day} \\[3pt]
\mathrm{TIR}(X_1) = \mathrm{TIR}(X_2) = 1
\end{array}
\label{eq:X}
\end{equation}

\noindent In our configuration, \( X_1 \) corresponds to Terra MODIS (1 km, 1 day), \( X_2 \) to Landsat 8 (30 m, 16 days), and \( X_3 \) to Sentinel-2 (10 m, 5 days), as summarized in Table~\ref{tab:Xconfiguration}. This formulation is general and can be extended to other RS missions that meet the conditions in Equation~\ref{eq:X}.

\begin{table}[ht]
\centering
\caption{Comparison of representative RS satellites with their thermal sensors, spatial and temporal resolutions.}
\renewcommand{\arraystretch}{1} 
\begin{tabular}[t]{@{\hspace{2pt}}l@{\hspace{6pt}}c@{\hspace{6pt}}c@{\hspace{6pt}}c@{\hspace{2pt}}}
\toprule
\textbf{Satellite} & \textbf{Thermal sensor} & \textbf{Spatial resolution} & \textbf{Temporal resolution} \\
\midrule
Terra       & MODIS  & 1 km   & 1 day   \\
Landsat 8   & TIRS   & 30 m   & 16 days \\
Sentinel-2  & /      & 10 m   & 5 days  \\
\bottomrule

\label{tab:Xconfiguration}
\end{tabular}
\end{table}

Given a target date \( t_2 \), the goal is to estimate LST at 10 m resolution, denoted as \( \hat{X}_3(s, t_2, LST, r_3) \), where \( r_3 = 10\,\text{m} \). To achieve this, we rely on the Terra MODIS LST at the same date \( X_1(s, t_2, LST, r_1) \) with \( r_1 = 1\,\text{km} \), as well as a prior reference date \( t_1 < t_2 \) where all RS satellites provide valid LST observations over the geographic region \( s \). At \( t_1 \), we construct a prior triple \( T_1 \) composed of:

\begin{itemize}
    \item Sentinel-2 spectral indices (\( r_3 = 10\,\text{m} \)): \( X_3(s, t_1, NDVI, r_3) \), \( X_3(s, t_1, NDWI, r_3) \), and \( X_3(s, t_1, NDBI, r_3) \).
    \item Landsat 8 spectral indices and LST (\( r_2 = 30\,\text{m} \)): \( X_2(s, t_1, NDVI, r_2) \), \( X_2(s, t_1, NDWI, r_2) \), \( X_2(s, t_1, NDBI, r_2) \),  and \( X_2(s, t_1, LST, r_2) \).
    \item Terra MODIS LST (\( r_1 = 1\,\text{km} \)): \( X_1(s, t_1, LST, r_1) \).
\end{itemize}

\noindent Here, NDVI, NDWI, and NDBI characterize vegetation, water, and built-up surfaces, respectively. Their formulations are summarized in Table~\ref{tab:indices}. 
These indices are widely used in LST studies due to their strong correlations with surface thermal behavior~\cite{guha2022land, abdalkadhum2021correlation}. 
We intentionally limit the indices to these three to maintain compactness and prevent redundancy.

Thus, given the prior triple \( T_1 \) and the Terra MODIS LST at \( t_2 \), the STF task is formulated as learning a DL network \( f \), parameterized by \( \mathbf{W} \), to estimate the target 10 m LST, as defined in Equation~\ref{eq:stf_dl}. Note that WGAST relies solely on a previous date $t_1$, and thus eliminates the need to wait for a future overlapping period.
\begin{equation}
\hat{X}_3(s, t_2, LST, r_3) = f\left(T_1, X_1(s, t_2, LST, r_1) \mid \mathbf{W}\right).
\label{eq:stf_dl}
\end{equation}

\begin{table}[ht]
\centering
\caption{Formulations of spectral indices for Landsat 8 and Sentinel-2. 
NIR: Near-Infrared, SWIR: Shortwave Infrared, Red and Green refer to visible bands. 
\( B_i \) denotes the \( i \)-th band of the corresponding satellite sensor.}
\renewcommand{\arraystretch}{1.1} 
\begin{tabular}[t]{@{\hspace{6pt}}l@{\hspace{16pt}}c@{\hspace{16pt}}c@{\hspace{16pt}}c@{\hspace{16pt}}}
\toprule
\textbf{Index} & \textbf{Equation} & \textbf{Landsat 8} & \textbf{Sentinel-2} \\
\midrule
NDVI  & $\frac{\text{NIR} - \text{Red}}{\text{NIR} + \text{Red}}$       & $\frac{B5 - B4}{B5 + B4}$   & $\frac{B8 - B4}{B8 + B4}$   \\
NDBI  & $\frac{\text{SWIR} - \text{NIR}}{\text{SWIR} + \text{NIR}}$     & $\frac{B6 - B5}{B6 + B5}$   & $\frac{B11 - B8}{B11 + B8}$ \\
NDWI  & $\frac{\text{Green} - \text{NIR}}{\text{Green} + \text{NIR}}$   & $\frac{B3 - B5}{B3 + B5}$   & $\frac{B3 - B8}{B3 + B8}$   \\
\bottomrule
\label{tab:indices}
\end{tabular}
\end{table}

\vspace{-1em}

\section{Methodology}

This section first reviews GANs and conditional GANs (cGANs), which form the basis of our approach. We then present the overall WGAST architecture and describe its main components: the generator, the weakly supervised learning strategy, the discriminator, and the loss functions.
\vspace{-0.5em}
\subsection{Generative Adversarial Networks}

GANs, introduced by~\cite{goodfellow2014generative}, are a powerful class of deep generative models. They are based on a two-player adversarial game between two networks, a generator \(G\) and a discriminator \(D\)~\cite{sharma2024generative}. The generator synthesizes realistic samples that mimic the distribution of real data \(x \sim p_{\text{data}}\), while the discriminator distinguishes them from generated samples \(G(z)\), where \(z \sim p_z\) is drawn from a latent prior distribution~\cite{hong2019generative}.

The discriminator outputs a probability map that indicates the likelihood that each input pixel is real or generated. It is optimized to maximize its classification accuracy, as expressed in Equation~\ref{eq:lossDF}.
\begin{equation}
\mathcal{L}_D = \max_D \; \mathbb{E}_{x \sim p_{\text{data}}}[\log D(x)] + \mathbb{E}_{z \sim p_z}[\log(1 - D(G(z)))]
\label{eq:lossDF}
\end{equation}

In contrast, the generator attempts to produce samples that fool the discriminator into classifying them as real. Its loss is defined in Equation~\ref{eq:lossGF}.
\begin{equation}
L^{(G)} = \min \left[ \log D(x) + \log (1 - D(G(z))) \right]
\label{eq:lossGF}
\end{equation}

The overall GAN training can be formulated as a minimax optimization problem as expressed in Equation \ref{eq:minmaxLoss}.
\begin{equation}
L = \min_G \max_D \left[ \log D(x) + \log (1 - D(G(z))) \right]
\label{eq:minmaxLoss}
\end{equation}

During training, the generator and discriminator engage in an adversarial interplay, as the generator improves at producing convincing samples, the discriminator enhances its ability to distinguish them from real ones. This iterative process drives both models to co-evolve and become increasingly competent.  Theoretically, this dynamic converges to a Nash equilibrium~\cite{hsieh2019finding}, where the generator has learned the true data distribution so well that the discriminator can no longer reliably differentiate real from fake.

cGANs~\cite{mirza2014conditional} extend standard GANs by providing additional input information, called the condition. This can be class labels, images, or other contextual data, which guide the generator to produce outputs dependent on it. As a result, cGANs enable more targeted and context-aware generation, valuable when outputs must align with a specific input modality.



\vspace{-0.75em}

\subsection{Overall Architeture}

\begin{figure}[h]
  \centering
  \includegraphics[width=0.5\textwidth]
  {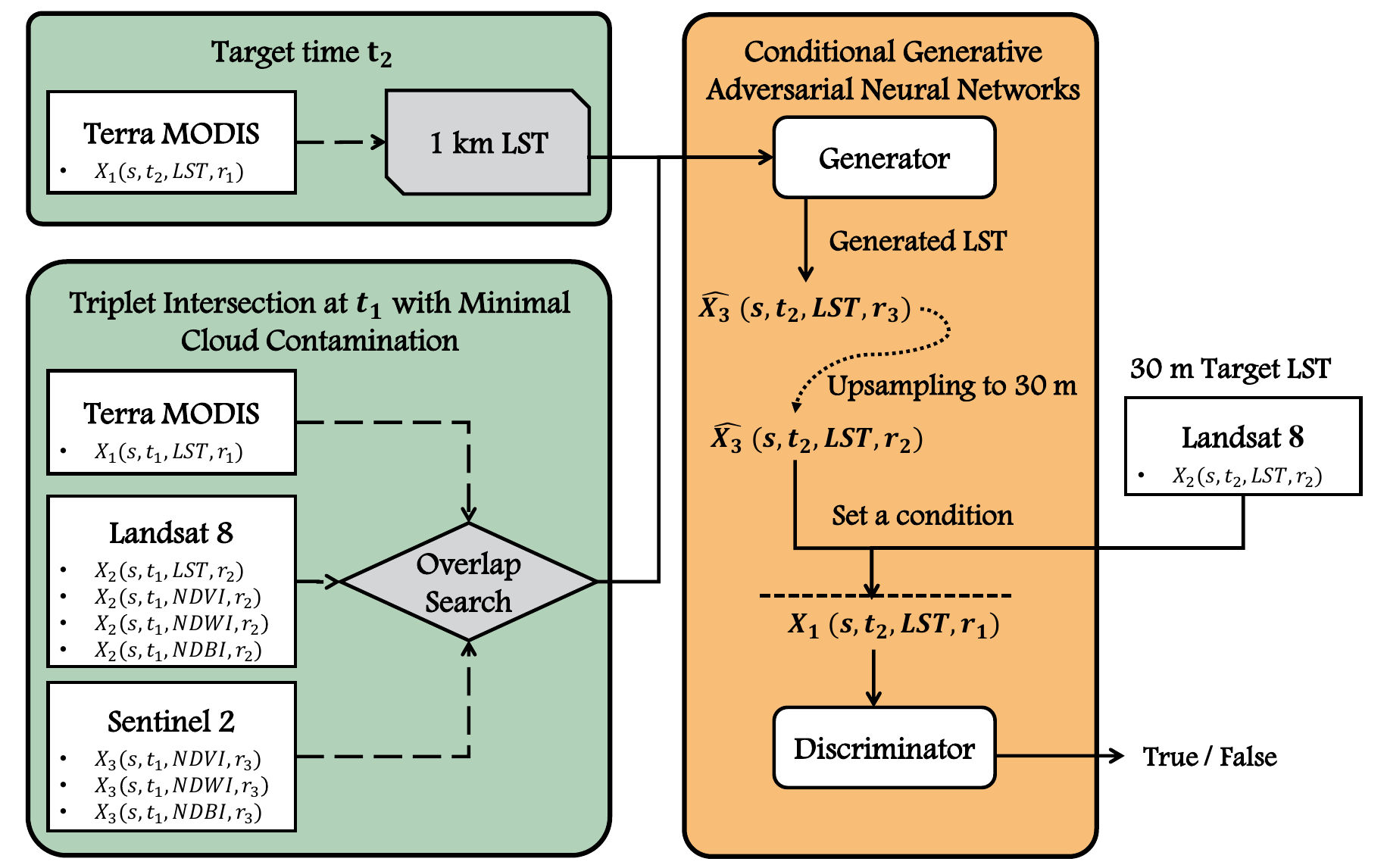}
    \caption{Overview of the WGAST framework. WGAST consists of a generator that fuses data from  Terra MODIS (\( r_1 = 1\,\text{km} \)), Landsat 8 (\( r_2 = 30\,\text{m} \)), and Sentinel-2 (\( r_3 = 10\,\text{m} \)) to produce a high-resolution LST estimate at 10\,m. A discriminator ensures the quality of the generated LST by comparing it with the reference Landsat 8 LST at 30\,m resolution.}

  \label{fig:architecture}
\end{figure}

The overall architecture of WGAST is shown in Fig.\ref{fig:architecture}. It is based on a cGAN, where generation is explicitly conditioned on the Terra MODIS LST observation at the target time. The goal is to estimate the LST at 10 m resolution for a given region \(s\) at target time \(t_2\). The generated output is denoted as \(\hat{X}_3(s, t_2, LST, r_3)\), where \(r_3 = 10\,\text{m}\). We chose a cGAN over other generative models for three reasons~\cite{he2025diffusion}. First, overlapping triples are limited, and cGANs can learn effectively from smaller datasets, whereas diffusion models usually require very large ones. Second, adversarial training ensures realistic, high-resolution LST patterns and enforces spatial consistency. Third, inference is efficient. Unlike diffusion models, which require iterative denoising, cGAN generates 10 m LST in a single forward pass, making it suitable for large-scale regions.

WGAST begins by identifying the overlapping timestep between the three RS satellites, as discussed in Section \ref{sec:problem_formulation}, to extract the triple \( T_1 \). This triple, along with the Terra MODIS LST observation at the target time \( t_2 \), \( X_1(s, t_2, LST, r_1) \) where \( r_1 = 1\,\text{km} \), is used as input to the generator module. The generator is designed to capture relevant spatio-spectral and spatio-temporal features from the inputs to produce a fine-resolution LST observation at 10 m for the target date $t_2$. Since no ground-truth LST data at 10 m resolution are available, we adopt a weakly-supervised strategy motivated by physical principles. Specifically, the generated LST is averaged using a \( 3 \times 3 \) window to approximate a 30 m resolution. This output is then fed to the discriminator, which receives both the downsampled generated LST and a reference Landsat 8 30 m LST, each conditioned on the Terra MODIS LST observation at time \( t_2 \). 
\vspace{-0.3em}

\begin{figure*}[h]
  \centering
  \includegraphics[width=1\textwidth]
  {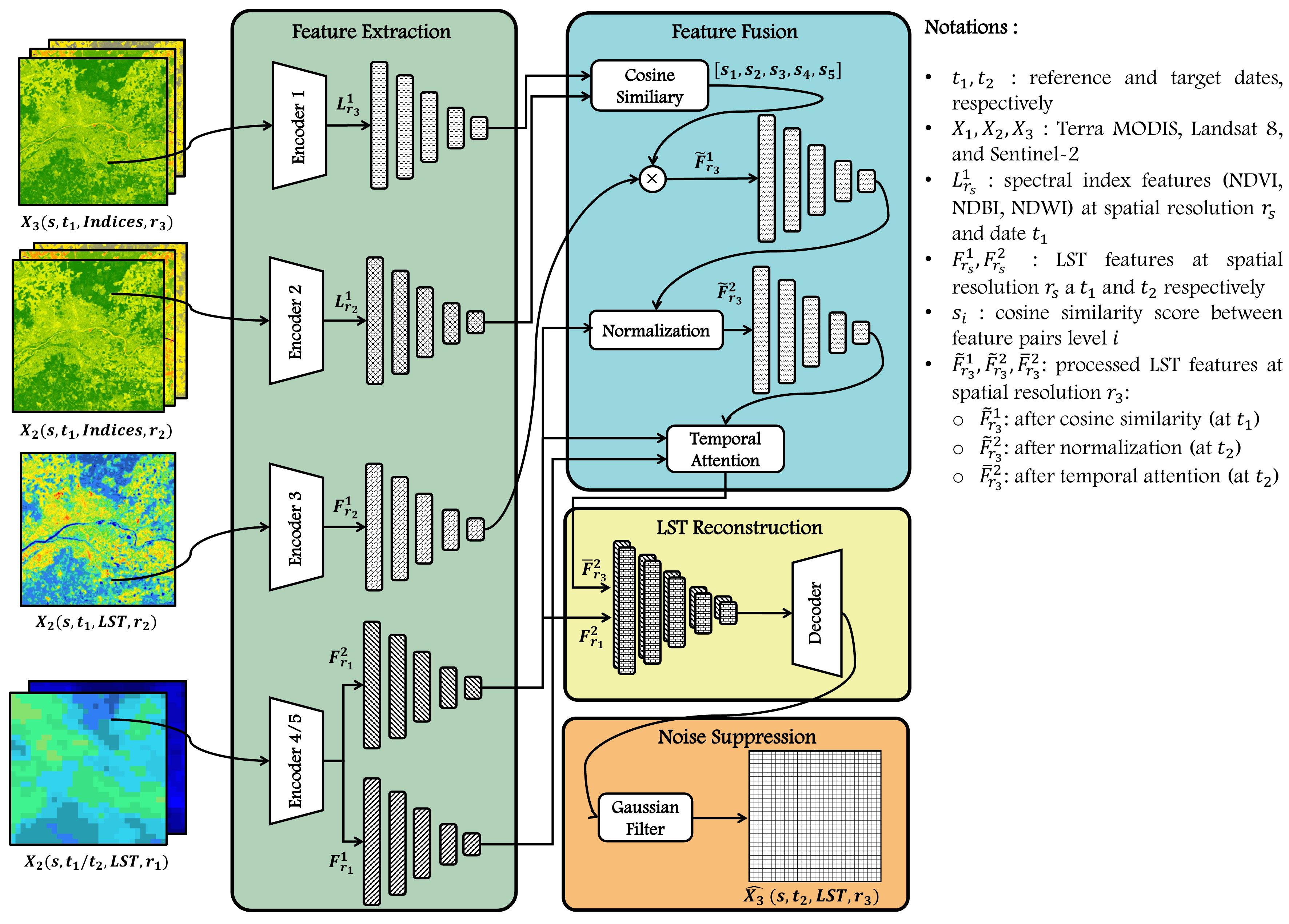}
    \caption{Overall architecture of the WGAST generator composed of four main stages: \textit{feature extraction}, \textit{feature fusion}, \textit{LST reconstruction}, and \textit{noise suppression}, with \( r_1 = 1\,\text{km} \), \( r_2 = 30\,\text{m} \), and \( r_3 = 10\,\text{m} \).}

  \label{fig:generator}
\end{figure*}

\subsection{Generator}
The WGAST generator, as depicted in Fig.\ref{fig:generator}, consists of four stages: \textit{feature extraction}, \textit{feature fusion}, \textit{LST reconstruction}, and \textit{noise suppression}.
\subsubsection{Feature extraction}
This stage uses a set of encoders composed of convolutional layers with downsampling and residual blocks to capture spatial and temporal features. Batch Normalization layers are omitted in the residual blocks, as they degrade GAN image synthesis quality~\cite{lim2017enhanced}. The extracted features are encoded into compact latent representations that enable the fusion stage to operate on the most informative abstractions. Specifically, five encoders produce multi-level feature sets as follows:

\begin{itemize}
  \item \textit{Encoder 1:} 10 m Sentinel-2 indices (NDVI, NDBI, NDWI) features at \( t_1 \), denoted as \( \mathbf{L}_{r_3}^{1} = \left\{ \mathbf{L}_{r_3,i}^{1} \right\}_{i=0}^{N} \).

  \item \textit{Encoder 2:} 30 m Landsat 8 indices (NDVI, NDBI, NDWI) features at \( t_1 \), denoted as \( \mathbf{L}_{r_2}^{1} = \left\{ \mathbf{L}_{r_2,i}^{1} \right\}_{i=0}^{N} \).

  \item \textit{Encoder 3:} 30 m Landsat 8 LST features at \( t_1 \),  denoted as \( \mathbf{F}_{r_2}^{1} = \left\{ \mathbf{F}_{r_2,i}^{1} \right\}_{i=0}^{N} \).

  \item \textit{Encoder 4:} 1 km Terra MODIS LST features at \( t_1 \), denoted as \( \mathbf{F}_{r_1}^{1} = \left\{ \mathbf{F}_{r_1,i}^{1} \right\}_{i=0}^{N} \).

  \item \textit{Encoder 5:} 1 km Terra MODIS LST features at \( t_2 \), denoted as \( \mathbf{F}_{r_1}^{2} = \left\{ \mathbf{F}_{r_1,i}^{2} \right\}_{i=0}^{N} \).
\end{itemize}

\smallskip

\noindent Here, \( N \) denotes the number of multi-level features produced by each encoder. Each feature group captures spatial or temporal characteristics at different levels of abstraction. Based on empirical evaluation, we fixed \( N = 5 \) to ensure a good balance between feature richness and computational efficiency.

\smallskip

\subsubsection{Feature fusion}
Feature fusion involves three operations: \textit{cosine similarity}, \textit{normalization}, and \textit{temporal attention}. The \textit{cosine similarity} between two feature vectors \( \mathbf{u}, \mathbf{v} \in \mathbb{R}^C \), where 
$C$ denotes the number of feature channels, is computed as in Equation~\ref{eq:cosine_similarity}.
\begin{equation}
\cos(\mathbf{u}, \mathbf{v}) = \frac{\mathbf{u} \cdot \mathbf{v}}{\|\mathbf{u}\|_2 \, \|\mathbf{v}\|_2}
\label{eq:cosine_similarity}
\end{equation}

\smallskip
\noindent Where \(\| \cdot \|_2\) denotes the L2 norm. This metric measures the cosine of the angle between two vectors, ranging from \(-1\) (opposite) to \(1\) (aligned), with 0 indicating no similarity. In image feature maps, it is applied element-wise across the feature space. In WGAST, The goal is to compute a similarity score between the multi-level features of Sentinel-2 indices at \( t_1 \), \( \mathbf{L}_{r_3}^{1} \), and the multi-level features of Landsat 8 indices at the same time,  \( \mathbf{L}_{r_2}^{1} \). This score is then used to spatially refine the Landsat 8 LST features \( \mathbf{F}_{r_2}^{1} \), to produce a 10 m features approximation of LST at \( t_1 \), denoted as \( \widetilde{\mathbf{F}}_{r_3}^{1} = \left\{ \widetilde{\mathbf{F}}_{r_3,i}^{1} \right\}_{i=0}^{N} \), where each level \( i \) is computed as in Equation \ref{eq:refined_features}.
\begin{equation}
\widetilde{\mathbf{F}}_{r_3,i}^{1} = \mathbf{F}_{r_2,i}^{1} \odot \text{cos}(\mathbf{L}_{r_2,i}^{1}, \mathbf{L}_{r_3,i}^{1})
\label{eq:refined_features}
\end{equation}


Unlike linear models that assume a global linear relationship between indices and LST values, cosine similarity is a non-linear measure that captures local, level-wise structural correspondences between feature maps at different resolutions. This allows the refinement process to emphasize features that are both semantically meaningful and spatially aligned, leading to more accurate and context-aware estimations.

In the second operation, we use \textit{Adaptive Instance Normalization} (AdaIN)~\cite{huang2017arbitrary} to harmonize the distributions between spatially detailed and temporally consistent features. Specifically, we align the 10 m LST features extracted from the pervious step \( \widetilde{\mathbf{F}}_{r_3}^{1} \) with the style statistics of Terra MODIS LST features at time \( t_2 \), \( \mathbf{F}_{r_1}^{2} \). AdaIN proceeds in two steps: each level \(i\) of \( \widetilde{\mathbf{F}}_{r_3}^{1} \) is first normalized by removing its mean and standard deviation, then rescaled and shifted using the statistics of the style features \( \mathbf{F}_{r_1}^{2} \). This is expressed in Equation~\ref{eq:adain}.


\vspace{-1em}
\begin{equation}
\begin{aligned}
\widetilde{\mathbf{F}}_{r_3}^{2} = \text{AdaIN}(\widetilde{\mathbf{F}}_{r_3}^{1}, \mathbf{F}_{r_1,i}^{2}) = \sigma(\mathbf{F}_{r_1,i}^{2}) \cdot \frac{\widetilde{\mathbf{F}}_{r_3}^{1} - \mu(\widetilde{\mathbf{F}}_{r_3}^{1})}{\sigma(\widetilde{\mathbf{F}}_{r_3}^{1})} \\ 
+ \mu(\mathbf{F}_{r_1,i}^{2}) \quad \quad \quad \quad \quad  \quad \quad \quad
\end{aligned}
\label{eq:adain}
\end{equation}

\noindent Where \( \mu(\cdot) \) and \( \sigma(\cdot) \) denote the mean and standard deviation of feature map level \( i \).

The final step, \textit{temporal attention}, models how temporal feature variations, \( \mathbf{F}_{r_1}^{1} \) and \( \mathbf{F}_{r_1}^{2} \), affect spatial feature structure. Its goal is to integrate this temporal information into the AdaIN-normalized spatial features \( \widetilde{\mathbf{F}}_{r_3}^{2} \). This is achieved through a spatially adaptive attention mechanism that estimates a set of attention weights \( \boldsymbol{\theta} = \left\{ \theta_i \right\}_{i=0}^{N} \), where each \( \theta_i \in [0, 1] \) represents the relative importance of temporal features at level \( i \). These weights are learned dynamically during training and control the influence of temporal variations in synthesizing the final 10 m feature representation at time \( t_2 \).  Specifically, a \(1 \times 1\) convolution with batch normalization projects the Terra MODIS LST features at \(t_1\) and \(t_2\) into a compact and discriminative space. Their difference captures temporal variations, which is processed through another \(1 \times 1\) convolution and a sigmoid activation to generate an attention mask. The mask then adaptively fuses the normalized spatial features with temporal features at \( t_2 \). The full procedure is detailed in Algorithm~\ref{alg:temporal_attention}.

\subsubsection{LST Reconstruction}
The reconstruction stage mirrors the encoder through a symmetric U-Net-like architecture composed of upsampling layers, transposed convolutions, and residual blocks to reconstruct the 10 m LST at time \( t_2 \). It takes as input the temporally attended spatial features \( \overline{F}_{r_3}^{2} \) and the original Terra MODIS LST features \( \mathbf{F}_{r_1}^{2} \), concatenated along the channel dimension. Reintroducing \( \mathbf{F}_{r_1}^{2} \) at this stage preserves large-scale temperature patterns that may have been weakened during earlier fusion and normalization. The combined features are decoded and progressively refined to produce the final 10 m LST output, denoted as \(\overline{X_3}(s, t_2, LST, r_3)\).

\subsubsection{Noise Suppresion}

The \textit{noise suppression} module reduces high-frequency artifacts present in the reconstructed LST \(\overline{X_3}(s, t_2, LST, r_3)\), to produce a spatially and temporally coherent result, \(\hat{X_3}(s, t_2, LST, r_3)\). This step is essential since LST fields in geophysical contexts exhibit smooth, spatially continuous patterns driven by heat diffusion rather than abrupt variations. However, deep generators with multiple upsampling layers and skip connections can unintentionally amplify pixel-level noise and produce unrealistic high-frequency textures that deviate from the expected physical smoothness of LST. To address this, we apply a Gaussian filter to smooth local fluctuations while preserving spatial structure and physical coherence. The Gaussian kernel is defined in Equation~\ref{eq:gaussian}.
\begin{equation}
G(x, y; \sigma) = \frac{1}{2\pi\sigma^2} \exp\left( -\frac{x^2 + y^2}{2\sigma^2} \right)
\label{eq:gaussian}
\end{equation}

\noindent where \( \sigma \) controls the smoothing degree. A higher \( \sigma \) produces stronger smoothing. In WGAST, \( \sigma \) is fixed to 1.


The noise suppression is then implemented via depthwise convolution of the predicted LST with the Gaussian kernel, as expressed in Equation~\ref{eq:smooth}.


\vspace{-1em}
\begin{equation}
\begin{aligned}
\hat{X_3}(s, t_2, LST, r_3) = \text{Conv2D}(\overline{X_3}(s, t_2, LST, r_3), \\
\; G(x, y; \tau))
\end{aligned}
\label{eq:smooth}
\end{equation}

\noindent where \( G(x, y; \tau) \) is the two-dimensional Gaussian kernel defined in Equation~\ref{eq:gaussian}, and the kernel size is \( k = 2 \cdot \lceil 3\sigma \rceil + 1 \). Reflective padding is used to reduce boundary artifacts.

\begin{algorithm}[t]
\KwIn{
    \( F_{r_1}^{1} \): Terra MODIS LST features at time \( t_1 \)\\
    \( F_{r_1}^{2} \): Terra MODIS LST features at time \( t_2 \)\\
    \( \widetilde{F}_{r_3}^{2} \): Normalized spatial features at time  \( t_2 \)\\
}

\KwOut{ \( \overline{F}_{r_3}^{2} \) : 10 m LST features at time \( t_2 \)}

$\overline{F}_{r_3}^{2} \gets [.] \cdot N$

\ForEach{i $= 1,…, N$}{ 
    
    $F1 \leftarrow \text{Conv1x1\_BN}(F_{r_1}^{1}[i])$

    $F2 \leftarrow \text{Conv1x1\_BN}(F_{r_1}^{2}[i])$

    $\Delta F \gets F1 - F2$

    $\theta_i \gets \text{Sigmoid}(\text{Conv1x1\_BN}(\Delta F))$
    
    $\overline{F}_{r_3}^{2}[i] \gets F_{r_1}^{2}[i] \cdot \theta_i + \widetilde{F}_{r_3}^{2}[i] \cdot (1 - \theta_i)$ 
}

\Return $\overline{F}_{r_3}^{2}$

\caption{Temporal attention-based fusion}
\label{alg:temporal_attention}
\end{algorithm}

\subsection{Weakly Supervised Learning}
In the absence of ground-truth LST data at 10 m resolution, we employ a weakly supervised learning strategy guided by physical principles. Prior studies have shown that coarse-resolution LST can be approximated by local averages of finer-resolution values~\cite{gao2016localization}. Under the assumption of limited thermal variability within small areas, a 30 m LST value can thus be estimated by averaging a $3 \times 3$ neighborhood of 10 m pixels. This corresponds to the inexact supervision paradigm of weakly supervised learning, where coarse-scale data are used to train models that predict fine-scale outputs~\cite{zhou2018brief}.

 
Based on this principle, we apply a $3 \times 3$ average pooling operation to the generator’s output $\hat{X_3}(s, t_2, LST, r_3)$ to obtain a 30 m estimate, as expressed in Equation~\ref{eq:agg}. This upsampled output is compared to the corresponding Landsat 8 LST observation at $t_2$, denoted $X_2(s, t_2, LST, r_2)$. The Landsat 8 LST is used only during training. This enables the model to depend solely on temporally frequent Terra MODIS inputs at inference time, thus preserving daily temporal coverage.


\vspace{-1em}
\begin{equation}
\begin{aligned}
\hat{X_3}(s, t_2, LST, r_2)(m, n) = \quad \quad \quad \quad \quad \quad \quad \quad \quad \quad \\
\frac{1}{9} \sum_{i=0}^{2} \sum_{j=0}^{2} \hat{X_3}(s, t_2, LST, r_3)(3m + i, 3n + j)
\end{aligned}
\label{eq:agg}
\end{equation}



\vspace{-0.5em}
\subsection{Discriminator}

\begin{figure}[h]
  \centering
  \includegraphics[width=0.5\textwidth]
  {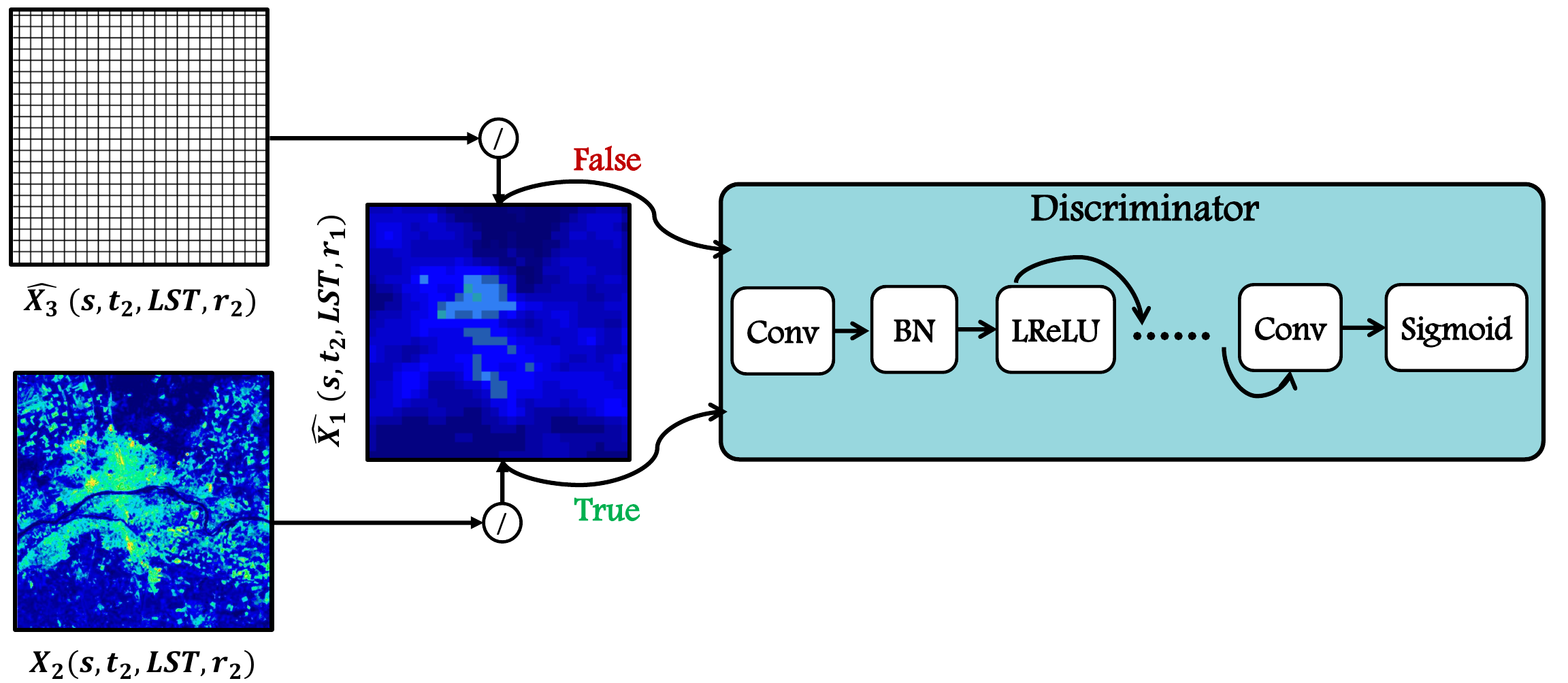}
\caption{WGAST discriminator based on the PatchGAN architecture.}

  \label{fig:discriminator}
\end{figure}

WGAST discriminator is built upon the PatchGAN architecture~\cite{goodfellow2014generative} (Fig.\ref{fig:discriminator}). Unlike standard GANs that classify entire images as real or fake, here it discriminates between observed and fused LST. The discriminator takes as input the fused LST conditioned on the corresponding Terra MODIS LST \( X_1(s, t_2, LST, r_1) \), which provides reliable thermal context at 1 km resolution. This conditioning enables the model to verify whether the target LST exhibits physically consistent temperature patterns with the Terra MODIS data. The final layer applies a softmax function to output probabilities, where higher values indicate greater confidence that a patch is real. During training, the discriminator is expected to output probabilities close to $1$ for real Landsat 8 LST samples \( X_2(s, t_2, LST, r_2) \) and close to $0$ for fused outputs \( \hat{X_3}(s, t_2, LST, r_2) \), both conditioned on the Terra MODIS input.


\vspace{-0.5em}
\subsection{Loss Functions}

\subsubsection{Discriminator Loss}

The discriminator is trained to distinguish between real LST observations and fused LST outputs generated by the model. Its objective function, denoted as $\mathcal{L}_D$, is defined in Equation~\ref{eq:disc_loss}. For brevity, we use the notation \(\hat{X}_3(LST, r_2)\) to represent \(\hat{X}_3(s, t_2, LST, r_2)\), and apply the same simplification to \(X_2\) and \(X_1\).

\begin{equation}
\begin{aligned}
\mathcal{L}_D = \frac{1}{2} \, \mathbb{E}_{\hat{X}_3, X_1} \left[ \left( D\left(\hat{X}_3(LST, r_2) \mid X_1(LST, r_1)\right) \right)^2 \right] \\
+ \frac{1}{2} \, \mathbb{E}_{X_2, X_1} \left[ \left( D\left(X_2(LST, r_2) \mid X_1(LST, r_1)\right) - 1 \right)^2 \right]
\end{aligned}
\label{eq:disc_loss}
\end{equation}

\noindent Where \( D(\cdot) \) denotes the discriminator. The loss adopts the least-squares GAN formulation~\cite{mao2017least}, which penalizes the discriminator for assigning values far from 0 for fused generated LST and far from 1 for real LST, thereby stabilizing training and mitigating vanishing gradients.  The expectations \(\mathbb{E}_{\cdot}\) denote averages taken over training samples.

\subsubsection{Generator Loss}

The generator loss, denoted by \(\mathcal{L}_G\), combines adversarial feedback with several image-based losses to ensure both realism and fidelity in the fused output. The full loss formulation is given in Equation~\ref{eq:gen_loss}.
\begin{equation}
\begin{aligned}
\mathcal{L}_G = \; & \alpha \, \underbrace{
\mathbb{E}_{\hat{X}_3, X_1} \left[ 
\left( D\left( \hat{X}_3(LST, r_2) \mid X_1(LST, r_1) \right) - 1 \right)^2 
\right]
}_{\mathcal{L}_{\text{GAN}}} \\ 
& + \beta \, \underbrace{
\frac{1}{K} \sum_{k=1}^K 
\left| \hat{X}_3(LST, r_2) - X_2(LST, r_2) \right|
}_{\mathcal{L}_{\text{content}}} \\
& + \gamma \, \underbrace{
\left( 
1 - \frac{ \langle \hat{X}_3(LST, r_2), X_2(LST, r_2) \rangle }
{ \| \hat{X}_3(LST, r_2) \|_2 \, \| X_2(LST, r_2) \|_2 }
\right)
}_{\mathcal{L}_{\text{spectrum}}} \\
& + \delta \, \underbrace{
\left(
1 - \text{MS-SSIM} \left( \hat{X}_3(LST, r_2), X_2(LST, r_2) \right)
\right)
}_{\mathcal{L}_{\text{vision}}}
\end{aligned}
\label{eq:gen_loss}
\end{equation}

\noindent Here, the adversarial loss $\mathcal{L}_{\text{GAN}}$ encourages the generator to produce fused LST outputs \(\hat{X}_3(s, t_2, LST, r_2)\) that the discriminator cannot distinguish from real LST observations, conditioned on the corresponding Terra MODIS input \(X_1(s, t_2, LST, r_1)\). The content loss \(\mathcal{L}_{\text{content}}\) is computed as the pixel-wise L1 distance, which promotes numerical consistency with real LST values. The spectrum loss \(\mathcal{L}_{\text{spectrum}}\), defined as one minus the cosine similarity, preserves the overall spatial variability and temperature distribution. Finally, the vision loss \(\mathcal{L}_{\text{vision}}\), derived from the Multiscale Structural Similarity Index (MS-SSIM), captures perceptual and structural coherence across multiple spatial scales. The hyperparameters $\alpha$, $\beta$, $\gamma$, and $\delta$ control the relative importance of each loss component in the overall generator loss function. In WGAST, we fixed $\alpha = 10^{-2}$, $\beta = 1$, $\gamma = 1$, and $\delta = 1$.

\section{Experimental Results}
\subsection{Region of Interest}

\begin{figure}[h]
  \centering
  \includegraphics[width=0.45\textwidth]
  {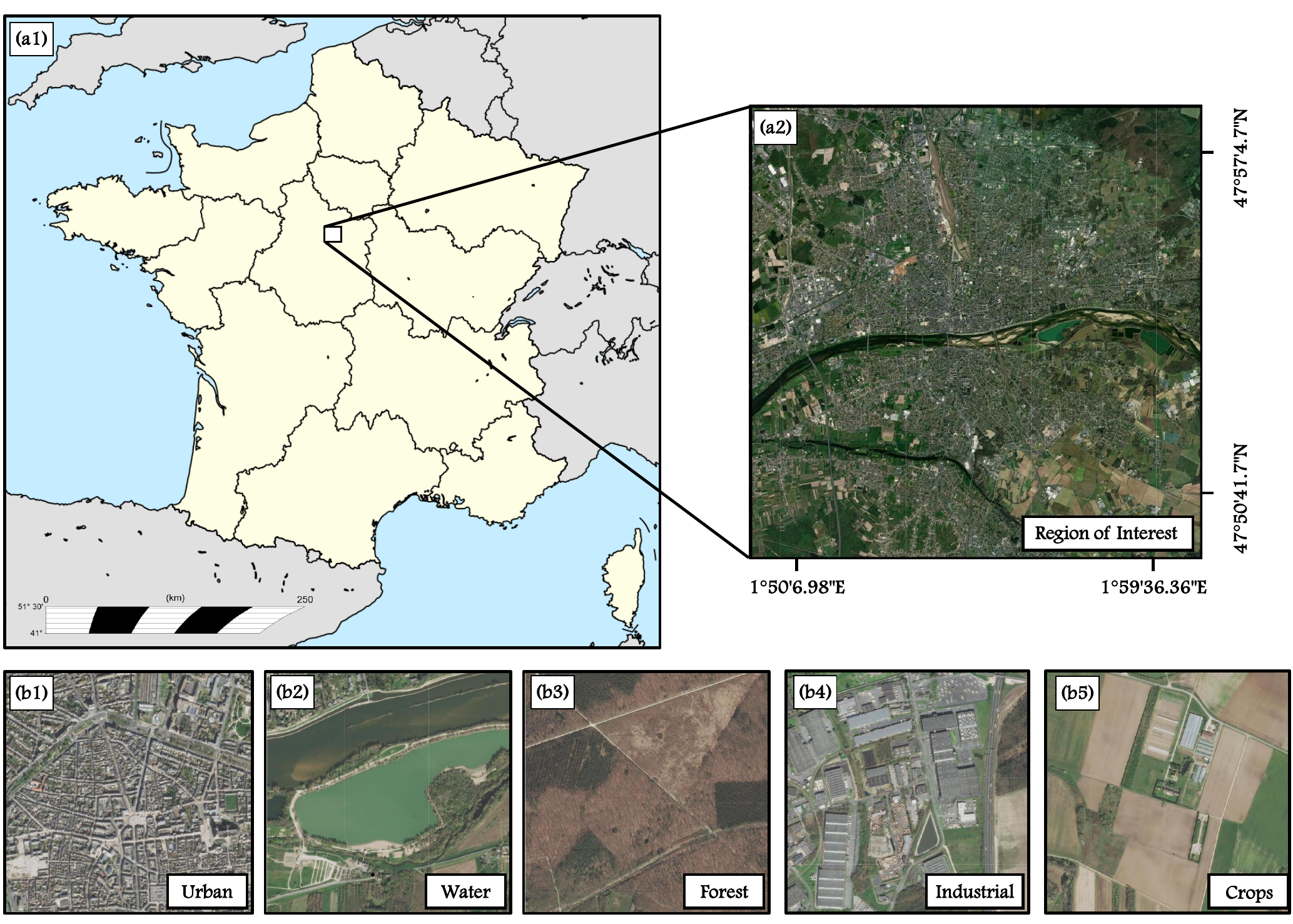}
    \caption{Geographic overview of the ROI. (a1) Position of the ROI within France. (a2) High-resolution satellite image of the ROI. (b1)–(b5) Representative examples of key land cover categories observed in the ROI: urban, water, forest, industrial, and agricultural zones.}
    
  \label{fig:ROI}
\end{figure}

The region of interest (ROI) is located within Orléans Métropole, France. As shown in Fig.\ref{fig:ROI}(a1), it spans latitudes 47$^\circ$50'41.77''N to 47$^\circ$54'1.74''N and longitudes 1$^\circ$50'6.98''E to 1$^\circ$59'36.36''E, covering approximately 114\,km$^2$. The Loire River, France’s longest, traverses the ROI and plays a significant geographical and thermal role (Fig.\ref{fig:ROI}(a2)). The area features a diverse land cover mosaic, including dense urban areas, water bodies, forests, industrial zones, and croplands (Fig.\ref{fig:ROI}(b1)–(b5)). Such heterogeneity introduces a wide range of radiative, thermal, and textural surface characteristics. This makes it a suitable benchmark for evaluating WGAST’s spatial adaptability and generalization, particularly in urban settings where LST varies sharply over short distances.

\subsection{Datasets}
\begin{table*}[h]
\centering
\caption{Summary of the 11 selected samples, each with a reference ($t_1$) and target ($t_2$) time.}
\renewcommand{\arraystretch}{1.1}
\begin{tabular}{c@{\hspace{10pt}}cccc@{\hspace{10pt}}ccc}
\toprule
\multirow{2}{*}{\textbf{Sample No.}} & \multicolumn{4}{c}{\textbf{Reference Date $t_1$}} & \multicolumn{3}{c}{\textbf{Target Date $t_2$}} \\
\cmidrule(lr){2-5} \cmidrule(lr){6-8}
& \textbf{Date} & \textbf{Terra MODIS} & \textbf{Landsat 8} & \textbf{Sentinel-2} & \textbf{Date} & \textbf{Terra MODIS} & \textbf{Landsat 8} \\
\midrule
1  & 09 Apr 2017 & 11:54 & 10:40 & 11:05 & 23 Feb 2018 & 11:54 & 10:40 \\
2  & 21 Oct 2018 & 11:54 & 10:41 & 11:06 & 26 Feb 2019 & 11:54 & 10:40 \\
3  & 06 Sep 2019 & 11:54 & 10:41 & 11:07 & 01 Apr 2020 & 11:54 & 10:40 \\
4  & 22 Jul 2020 & 11:54 & 10:41 & 11:07 & 07 Aug 2020 & 11:54 & 10:41 \\
5  & 06 Mar 2022 & 11:48 & 10:41 & 10:57 & 22 Mar 2022 & 11:48 & 10:41 \\
6  & 13 Aug 2022 & 11:42 & 10:41 & 10:57 & 29 Aug 2022 & 11:42 & 10:41 \\
7  & 28 May 2023 & 11:10 & 10:40 & 11:07 & 13 Jun 2023 & 10:36 & 10:40 \\
8  & 12 Apr 2024 & 10:35 & 10:40 & 11:07 & 19 Sep 2024 & 10:00 & 10:41 \\
9  & 19 Sep 2024 & 10:00 & 10:41 & 11:07 & 05 Oct 2024 & 10:48 & 10:41 \\
10 & 19 Sep 2024 & 10:00 & 10:41 & 11:07 & 21 Oct 2024 & 10:06 & 10:41 \\
11 & 19 Sep 2024 & 10:00 & 10:41 & 11:07 & 01 May 2025 & 09:30 & 10:40 \\
\bottomrule
\end{tabular}
\label{tab:satellite_samples_header}
\end{table*}

\subsubsection{Satellite Data}
The RS satellite data $X_1$, $X_2$, and $X_3$ were obtained from the Google Earth Engine (GEE) platform~\cite{gorelick2017google}. Terra MODIS data were accessed from the MOD11A1 (Collection 6.1) daily 1 km product, which provides atmospherically corrected LST values via the \textit{LST\_Day\_1km} band, with a reported RMSE within 2\,$^\circ$C~\cite{duan2019validation}. Landsat 8 data were obtained from the USGS Level-2 Collection 2 Tier 1 dataset, with LST derived from the \textit{ST\_B10} thermal band and estimated using a single-channel algorithm, with an accuracy of approximately $1.5\,^\circ$C~\cite{jimenez2014land}. NDVI, NDWI, and NDBI were computed from SR bands \textit{SR\_B3}, \textit{SR\_B4}, \textit{SR\_B5}, and \textit{SR\_B6}. Sentinel-2 imagery was collected from the Harmonized MSI Level-2A collection, with NDVI, NDWI, and NDBI computed from SR bands \textit{B2}, \textit{B3}, \textit{B4}, \textit{B8}, and \textit{B11}. Table~\ref{tab:satellite_samples_header} lists the $11$ selected samples, each consisting of a reference date \( t_1 \) and target date \( t_2 \). At \( t_1 \), Terra MODIS, Landsat 8, and Sentinel-2 acquisitions overlapped spatially and temporally. At \( t_2 \), WGAST uses only Terra MODIS LST for input, with Landsat 8 reserved for evaluation. For example, to predict 10 m LST on 23 Feb 2018 (\(t_2\)), WGAST takes the Terra MODIS LST at \(t_2\) and the triple \(T_1\) from 09 Apr 2017 (\(t_1\)) to generate a fused LST, which is then upsampled and compared with Landsat 8 acquired on \(t_2\). The satellite overpasses are closely aligned, with a maximum time difference of about 1 h 15 min, occurring between 10:00 and 12:00 local time. This mid-morning window ensures stable surface and atmospheric conditions, and thus minimizes variations in LST and SR, and reinforces the suitability of the fusion process. Resolutions are $1200 \times 1200$ pixels for Sentinel-2 and $400 \times 400$ for Landsat 8, with all samples having cloud contamination below 20\%. Missing pixels due to clouds or technical issues were filled via adaptive spatial interpolation using a $3 \times 3$ local mean window, expanded progressively by steps of 2 until at least one valid neighbor was included.

The first 7 samples were used for training, with patch sizes of $96 \times 96$ and a stride of $24$ for Sentinel-2 ($32 \times 32$ and a stride of $8$ for Landsat 8), yielding 15,463 training patches. The remaining 4 samples were reserved for testing. To prevent temporal leakage, reference dates $t_1$ of a sample were not allowed to coincide with target dates $t_2$ of other training samples. Testing samples were independent, thus overlap was allowed. WGAST was trained with a learning rate of $2 \times 10^{-4}$ and batch size of 32 on an NVIDIA RTX A6000 GPU.

\subsubsection{In-situ Measurement}
\begin{figure}[h]
  \centering
  \includegraphics[width=0.46\textwidth]
  {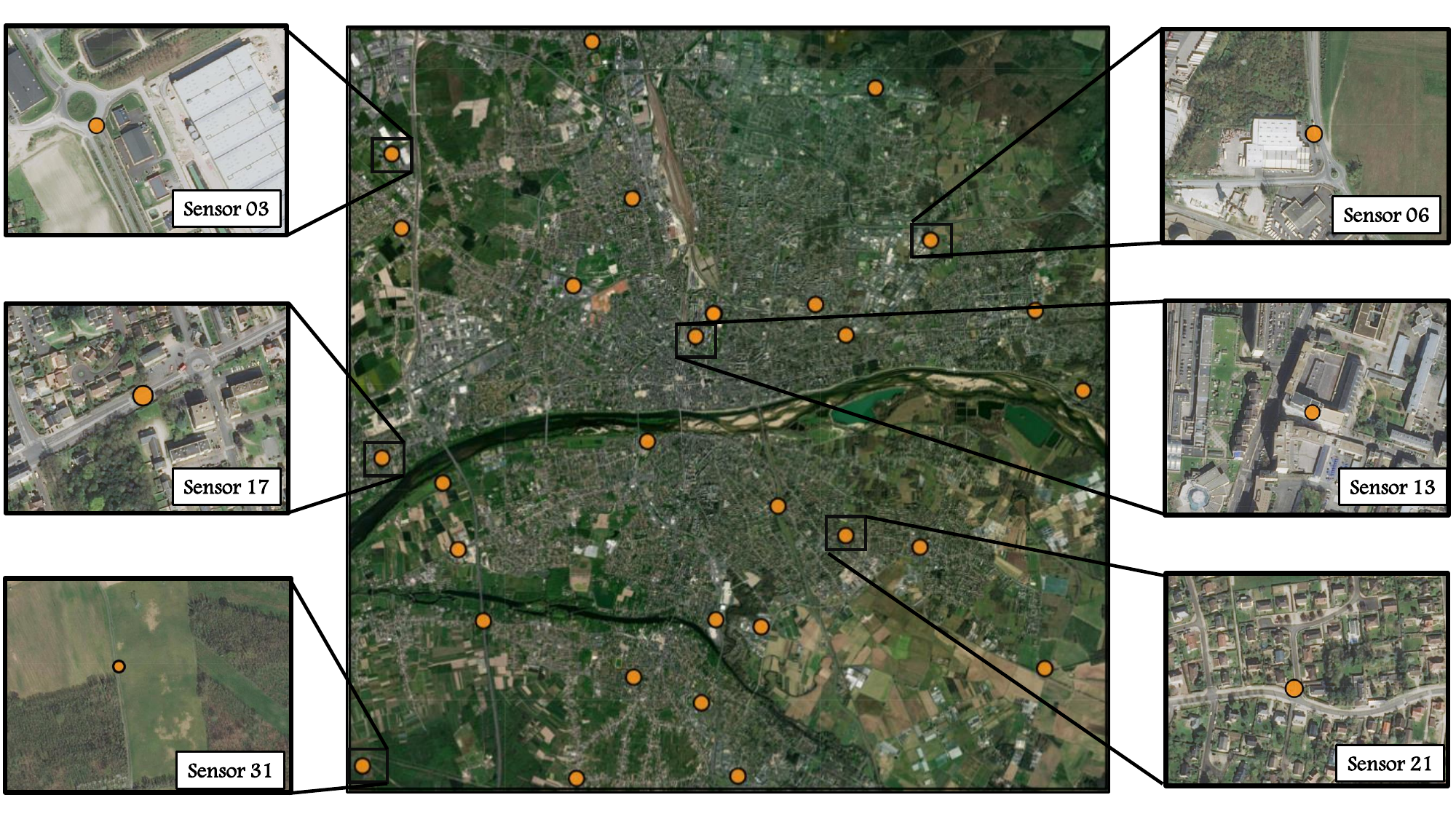}

\caption{Spatial distribution of the in-situ near-ground sensors across the ROI. The central map shows all sensor locations (orange dots), while the zoomed-in insets highlight the detailed placement of selected sensors (Sensors 03, 06, 13, 17, 21, and 31) within their immediate surroundings.}

  \label{fig:sensors}
\end{figure}

To validate WGAST under real-world conditions, we used a network of 33 near-ground sensors distributed across the ROI, each measuring near-surface air temperature ($T_a$) at a height of 2 m following standard meteorological protocols. The validation period extended from 1 April to 30 May 2025, during which 25 cloud-free Terra MODIS LST images were acquired. Accordingly, WGAST generated 25 cloud-free LST maps at 10 m resolution for the same dates. Fig.\ref{fig:sensors} shows the spatial distribution of the sensors. During this period, only two Landsat 8 LST acquisitions (1 May and 17 May 2025) were available, both partially affected by clouds. Moreover, no usable Landsat 8 data were available in April due to persistent cloud cover and technical issues. In contrast, WGAST produced $25$ complete LST observations over the same timeframe.

Although $T_a$ and LST represent distinct physical variables, they exhibit strong radiative coupling under clear-sky daytime conditions. This relationship makes $T_a$ a reliable proxy for indirectly evaluating the consistency and physical realism of WGAST’s fused LST outputs. Additionally, field observations indicate that $T_a$ remains relatively stable within a 10 m spatial window, making it comparable to the 10 m WGAST-derived LST maps. Thus, for each sensor location, we extracted its exact geographic coordinates from the fused LST maps and assigned the corresponding 10 m\,$\times$\,10 m pixel value as the representative LST. This produced a spatially aligned dataset of paired $T_a$ and LST values.

\subsection{Quantitative Assessment}

\subsubsection{Convergence Analysis}
\begin{figure}[h]
  \centering
  \includegraphics[width=0.45\textwidth]
  {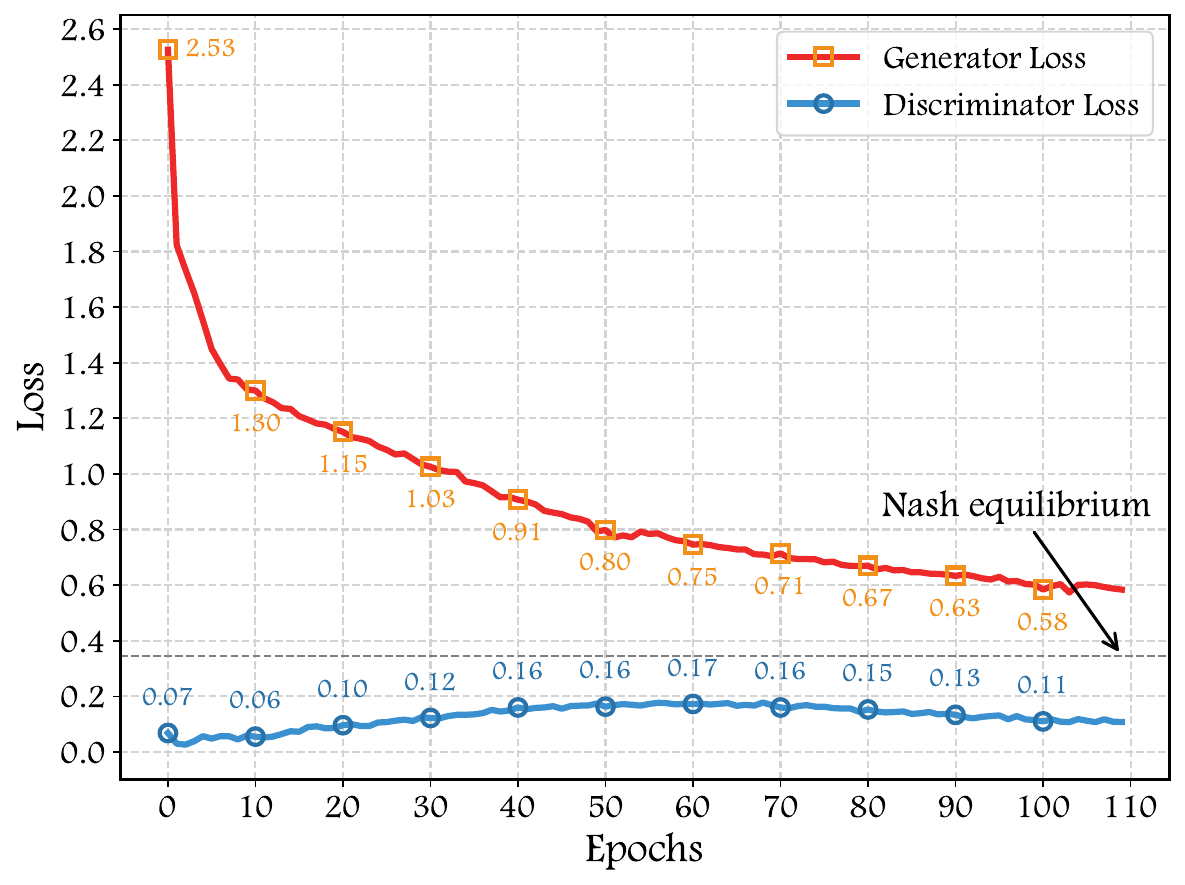}
\caption{Evolution of generator and discriminator losses during WGAST training. The generator loss decreases as it learns to produce more realistic fused LST, while the discriminator loss increases as its task becomes more challenging. Both losses eventually stabilize.}
  \label{fig:training_loss}
\end{figure}

Fig.\ref{fig:training_loss} shows the training evolution of the generator and discriminator losses during adversarial training. The generator loss starts relatively high at $\sim 2.53$, reflecting its initial inability to capture complex spatio-temporal relationships. As training progresses and the generator produces more realistic fused LST, the loss decreases steadily to $\sim 0.58$. Conversely, the discriminator begins with a low loss of $\sim 0.07$, indicating that distinguishing real from generated LST is initially easy due to the low quality of fused outputs. As the generator improves, the discriminator’s task becomes more difficult, leading to a gradual increase in its loss. This opposing behavior illustrates the adversarial dynamics of cGANs. Eventually, both losses stabilize near a Nash equilibrium, which indicates convergence.

\begin{table*}[h]
\centering
\caption{Quantitative comparison of WGAST against BicubicI, Ten-ST~\cite{mhawej2022daily}, GAN-10, AE-10 and FuseTen~\cite{bouaziz2025fuseten} over four dates.}
\renewcommand{\arraystretch}{1}
\small

\setlength{\tabcolsep}{3pt} 
\begin{tabular}{lcccccccccccc}
\toprule
\textbf{Metric} & \textbf{BicubicI} & \textbf{Ten-ST} & \textbf{GAN-10} & \textbf{AE-10} &  \textbf{FuseTen} & \textbf{WGAST} &
\textbf{BicubicI} & \textbf{Ten-ST} & \textbf{GAN-10} & \textbf{AE-10} & \textbf{FuseTen} & \textbf{WGAST} \\
\cmidrule(lr){2-7} \cmidrule(lr){8-13}

 & \multicolumn{6}{c}{\textbf{19 Sep 2024}} & \multicolumn{6}{c}{\textbf{05 Oct 2024}} \\
\midrule

RMSE ($\downarrow$)   & 3.637 & 3.934 & 4.589 & 3.396 & 3.220 & \textbf{2.619} & 2.451 & 2.583 & 2.008& 1.623 & \textbf{1.050} & 1.386 \\
SSIM ($\uparrow$)     & 0.640 & 0.526 & 0.365 & 0.424 & 0.798 & \textbf{0.814} & 0.666 & 0.621 & 0.491& 0.553 & 0.863 & \textbf{0.912} \\
PSNR ($\uparrow$)     & 15.357 & 14.503 & 13.826 & 15.953 & 18.552 & \textbf{21.399} & 17.765 & 17.310 &19.497& 21.348 &  \textbf{25.657} & 24.990 \\
SAM ($\downarrow$)    & 4.841 & 6.168 & 6.143 & 5.509 & \textbf{3.865} & 5.428 & 3.162 &4.677& 6.158 & 4.658 & \textbf{2.823} & 3.604 \\
CC ($\uparrow$)       & 0.572 & -0.025 & 0.431 & 0.437 & \textbf{0.814} & 0.747 & 0.571 & 0.113 & 0.664& 0.663 &  \textbf{0.901} & 0.860 \\
ERGAS ($\downarrow$)  & 4.595 & 4.983 & 5.798 & 4.290 & 3.944 & \textbf{3.309} & 4.190 & 4.447 & 3.433& 2.774 &  \textbf{1.795} & 2.293 \\

\midrule

 & \multicolumn{6}{c}{\textbf{21 Oct 2024}} & \multicolumn{6}{c}{\textbf{01 May 2025}} \\
 
\midrule

RMSE ($\downarrow$)   & 2.150 & 2.342 & 2.676& 2.294 & 1.905 & \textbf{1.487} & 4.724 & 5.340 &6.198& 3.875 & 3.234 & \textbf{2.312} \\
SSIM ($\uparrow$)     & 0.770 & 0.837 & 0.414& 0.519 & 0.867 & \textbf{0.895} & 0.538 & 0.707 & 0.486 & 0.582 & 0.790 & \textbf{0.841} \\
PSNR ($\uparrow$)     & 18.450 & 24.592 & 16.582 &  17.888 & 21.748 & \textbf{25.322} & 16.186 & 21.281 & 13.827 & 17.906 & 19.476 & \textbf{23.105} \\
SAM ($\downarrow$)    & 3.930 & 4.689 & 4.690& 3.813 & \textbf{3.033} & 3.673 & 5.823 & 7.625 & 5.782& 5.739 & 3.918 & \textbf{3.888} \\
CC ($\uparrow$)       & 0.494 & 0.120 &0.546& 0.579 & \textbf{0.896} & 0.887 & 0.538 & 0.095 & 0.556& 0.580 & \textbf{0.840} & 0.826 \\
ERGAS ($\downarrow$)  & 3.315 & 3.610 & 4.126& 3.537 & 2.937 & \textbf{2.293} & 4.819 & 5.515& 6.323 & 3.953 & 3.300 & \textbf{2.358} \\
\bottomrule
\end{tabular}

\label{tab:quantitive}
\end{table*}

\smallskip

\subsubsection{Model Performance Analysis}
WGAST was evaluated over four dates using six established metrics. Two are error metrics: Root Mean Square Error (RMSE) and Error Relative Global Dimensionless Synthesis (ERGAS), which quantify pixel-wise discrepancies between the generated and reference LST. The remaining four are quality metrics: Structural Similarity Index Measure (SSIM), Peak Signal-to-Noise Ratio (PSNR), Spectral Angle Mapper (SAM), and Correlation Coefficient (CC), which assess perceptual, spectral, and structural fidelity. For evaluation, predicted 10 m LSTs were averaged within a $3 \times 3$ window to match the 30 m resolution and validated against Landsat 8 LST. Each test image was divided into overlapping patches, and WGAST predicted LST for each patch independently. Overlapping predictions were then merged using soft border weighting to reduce boundary artifacts and ensure smooth reconstruction. We compared WGAST against five baselines. The first three are BicubicI, Ten-ST~\cite{mhawej2022daily}, and FuseTen~\cite{bouaziz2025fuseten}. BicubicI performs bicubic interpolation to upscale Terra MODIS LST to Sentinel-2 resolution, and thus produces daily LST at 10 m. Ten-ST assumes a linear relationship and applies robust least-squares fusion of Terra MODIS and Sentinel-2 data to generate daily 10 m LST. FuseTen employs a hybrid cGAN model, with a generator integrating a linear regression component. As fully DL-based methods for daily 10 m LST estimation are currently unavailable, we introduce two additional baselines, GAN-10 and AE-10. These first estimate daily 30 m LST using a GAN~\cite{song2022mlff} or AutoEncoder~\cite{tan2019enhanced}, then downsample to 10 m using linear regression guided by Sentinel-2 spectral indices from the prior date $t_1$.

On 19~Sep~2024, WGAST outperforms most competing methods, except for SAM and CC. Compared to FuseTen, it reduces RMSE by $18.66$\%, increases SSIM by $2.01$\%, improves PSNR by $15.35$\%, and decreases ERGAS by $16.10$\%, indicating more accurate pixel-wise LST estimates and enhanced spatial and spectral fidelity. The slightly higher SAM and marginally lower CC reflect WGAST’s emphasis on preserving local spatial details and structural sharpness, occasionally at the expense of global angular alignment. Compared to Ten-ST-GEE, WGAST achieves even larger gains, including a $33.43$\% RMSE reduction, $54.75$\% SSIM increase, $47.55$\% PSNR gain, and $33.60$\% ERGAS decrease. Relative to AE-10, RMSE drops by $22.88$\% and SSIM increases by $91.98$\%, with similar improvements across other metrics.  On 05~Oct~2024, WGAST performance is comparable to FuseTen, with an RMSE of $1.386$, slightly higher than FuseTen’s $1.050$. Similar trends are observed for PSNR, SAM, CC, and ERGAS, while SSIM improves by $5.68$\%. This is likely due to the LST on this day exhibiting a linear spatio-temporal trend, which allowed FuseTen to better capture the absolute values. On 21~Oct~2024, WGAST delivers strong performance across all metrics. Compared to FuseTen, it reduces RMSE by $21.94$\%, increases SSIM by $3.23$\%, gains $16.43$\% in PSNR, and lowers ERGAS by $21.93$\%. Compared to Ten-ST-GEE, RMSE decreases by $36.51$\%, SSIM improves by $6.93$\%, PSNR increases by $2.97$\%, and ERGAS drops by $36.49$\%. Relative to AE-10, WGAST reduces RMSE by $35.18$\% and improves SSIM by $72.45$\%. On 01~May~2025, despite the large temporal gap from the reference date at $t_1$ (19~Sep~2024), WGAST maintains strong performance, whereas other methods show notable accuracy degradation. For instance, compared to FuseTen, WGAST achieves a $28.51$\% RMSE reduction and a $6.46$\% SSIM gain.

On average across all test dates, as shown in Table~\ref{tab:average}, WGAST reduces RMSE by $17.05$\%, increases SSIM by $4.22$\%, improves PSNR by $10.98$\%, and lowers ERGAS by $13.76$\% compared to FuseTen. Relative to the best linear method, the gains are even more pronounced, with a $39.80$\% RMSE reduction, $28.53$\% SSIM increase, $22.05$\% PSNR gain, and $44.41$\% ERGAS decrease, demonstrating consistent superiority across all evaluation metrics. These results indicate that linear baselines, hybrid models, or combinations of STF and downscaling approaches do not achieve better performance than a fully DL–based model.

\setlength{\tabcolsep}{2.5pt} 
\begin{table}[h]
\centering
\caption{Average quantitative comparison of all methods.}
\renewcommand{\arraystretch}{1}
\small

\begin{tabular}{l@{\hspace{-5pt}}cccccc}
\toprule
\textbf{Metric} & \textbf{BicubicI} & \textbf{Ten-ST}& \textbf{GAN-10} & \textbf{AE-10} & \textbf{FuseTen} & \textbf{WGAST}\\
\cmidrule(lr){2-7}

& \multicolumn{6}{c}{\textbf{Average}}\\
\midrule
RMSE($\downarrow$)   & 3.241 & 3.550 & 3.868 & 2.637 & 2.352 & \textbf{1.951} \\
SSIM($\uparrow$)     & 0.654 & 0.673 &0.439&0.520& 0.830 & \textbf{0.865}\\
PSNR($\uparrow$)     & 16.940 & 19.422 &15.933&  18.274 & 21.358 & \textbf{23.704}  \\
SAM($\downarrow$)    & 4.439 & 6.160 & 5.693 & 4.930& \textbf{3.410} & 4.148 \\
CC($\uparrow$)       & 0.544 & 0.076 & 0.549& 0.565& \textbf{0.863} & 0.830  \\
ERGAS($\downarrow$)  & 4.073 & 4.645 &  4.920 & 3.639& 2.994 & \textbf{2.582} \\
\bottomrule
\end{tabular}
\label{tab:average}
\end{table}

\smallskip
\subsubsection{In-situ Measurement}
LST and near-surface air temperature (\textit{T\textsubscript{a}}) are fundamentally distinct physical quantities and are not directly comparable in a quantitative sense. However, under clear-sky and daytime conditions, they exhibit strong radiative coupling due to energy exchanges between the land surface and the atmosphere, which often leads to coherent spatial and temporal patterns. Accordingly, while the absolute values of fused 10 m LST and \(T_a\) may differ, their monotonic behavior is typically consistent within specific seasons~\cite{naserikia2023land}. In our study, we focus on the period from 01 April to 01 June 2025, which corresponds to a single season, thus allowing this assumption to be reasonably validated.  Therefore, a high correlation between fused LST and \(T_a\) serves as a meaningful indicator of the physical realism and internal consistency of the predicted LST. We used the Pearson correlation coefficient (PCC) and Spearman rank correlation coefficient (SRCC) to quantify this relationship. PCC measures the strength and direction of a linear relationship between two continuous variables~\cite{cohen2009pearson}, whereas SRCC quantifies the strength and direction of a monotonic relationship~\cite{zar2014spearman}, making it suitable for non-linear monotonic associations. Both metrics range from \(-1\) to \(1\), with values near either extreme indicating strong linear or monotonic relationships, respectively. Given two variables $a$ and $b$ with ranks $R(u)$ and $R(v)$, the two metrics are defined in Equation~\ref{eq:correlation}.
\begin{equation}
\text{PCC} = \frac{\mathrm{cov}(a, b)}{\sigma_a \sigma_b}, \quad
\text{SRCC} = 1 - \frac{6 \sum_{i=1}^m d_i^2}{m(m^2 - 1)}
\label{eq:correlation}
\end{equation}

\noindent where \( \mathrm{cov}(a, b) \) denotes the covariance between \( a \) and \( b \), \( \sigma_a \) and \( \sigma_b \) are their standard deviations, \(d_i = R(a_i) - R(b_i)\) is the difference between the ranks of the \(i\)-th values of the two variables, and $m$ is the total number of paired data. 

Table \ref{tab:lstsat} reports the PCC and SRCC between the WGAST's 10 m pixel-wise LST and the corresponding near-surface air temperature (\(T_a\)) across 33 sensors. The results reveal consistently strong correlations, with PCC values ranging from $0.80$ to $0.95$ and SRCC values from $0.80$ to $0.94$. These indicate a strong and statistically meaningful relationship between the generated LST and the near-surface \(T_a\), which supports the physical realism and consistency of WGAST's 10 m LST. Notably, even the lowest values exceed commonly accepted thresholds, highlighting the method’s robustness across diverse sensor locations and urban conditions. Overall, this performance underscores WGAST’s ability to reliably capture spatio-temporal variations in LST at fine spatial resolution.

\begin{table}[ht]
\centering
\caption{PCC and SRCC between fused LST and near-surface air temperature $T_a$ across sensors}
\setlength{\tabcolsep}{4pt} 
\renewcommand{\arraystretch}{0.5}
{\small
\begin{tabular}{l *{9}{c}}
\toprule
\textbf{Sensor} & 1 & 2 & 3 & 4 & 5 & 6 & 7 & 8 & 9 \\
\midrule
\textbf{PCC} & 0.93 & 0.92 & 0.93 & 0.90 & 0.95 & 0.91 & 0.91 & 0.88 & 0.88 \\
\textbf{SRCC} & 0.94 & 0.93 & 0.89 & 0.85 & 0.94 & 0.91 & 0.88 & 0.85 & 0.87 \\
\bottomrule
\end{tabular}

\vspace{0.5em}

\begin{tabular}{l *{9}{c}}
\toprule
\textbf{Sensor} & 10 & 11 & 12 & 13 & 14 & 15 & 16 & 17 & 18 \\
\midrule
\textbf{PCC} & 0.80 & 0.89 & 0.90 & 0.91 & 0.90 & 0.91 & 0.90 & 0.95 & 0.91 \\
\textbf{SRCC} & 0.80 & 0.88 & 0.88 & 0.93 & 0.90 & 0.92 & 0.87 & 0.93 & 0.94 \\
\bottomrule
\end{tabular}

\vspace{0.5em}

\begin{tabular}{l *{9}{c}}
\toprule
\textbf{Sensor} & 19 & 20 & 21 & 22 & 23 & 24 & 25 & 26 & 27 \\
\midrule
\textbf{PCC} & 0.88 & 0.89 & 0.88 & 0.90 & 0.90 & 0.86 & 0.91 & 0.88 & 0.88 \\
\textbf{SRCC} & 0.85 & 0.86 & 0.85 & 0.87 & 0.88 & 0.88 & 0.91 & 0.85 & 0.85 \\
\bottomrule
\end{tabular}

\vspace{1em}

\begin{tabular}{l *{6}{cc}}
\toprule
\textbf{Sensor} & 28 & 29 & 30 & 31 & 32 & 33 & Average \\
\midrule
\textbf{PCC} & 0.93 & 0.90 & 0.90 & 0.91 & 0.89 & 0.93 & 0.90 \\
\textbf{SRCC} & 0.91 & 0.85 & 0.87 & 0.87 & 0.91 & 0.93 & 0.89 \\
\bottomrule
\end{tabular}
}
\label{tab:lstsat}
\end{table}

\begin{figure}[h]
  \centering
  \includegraphics[width=0.5\textwidth]
  {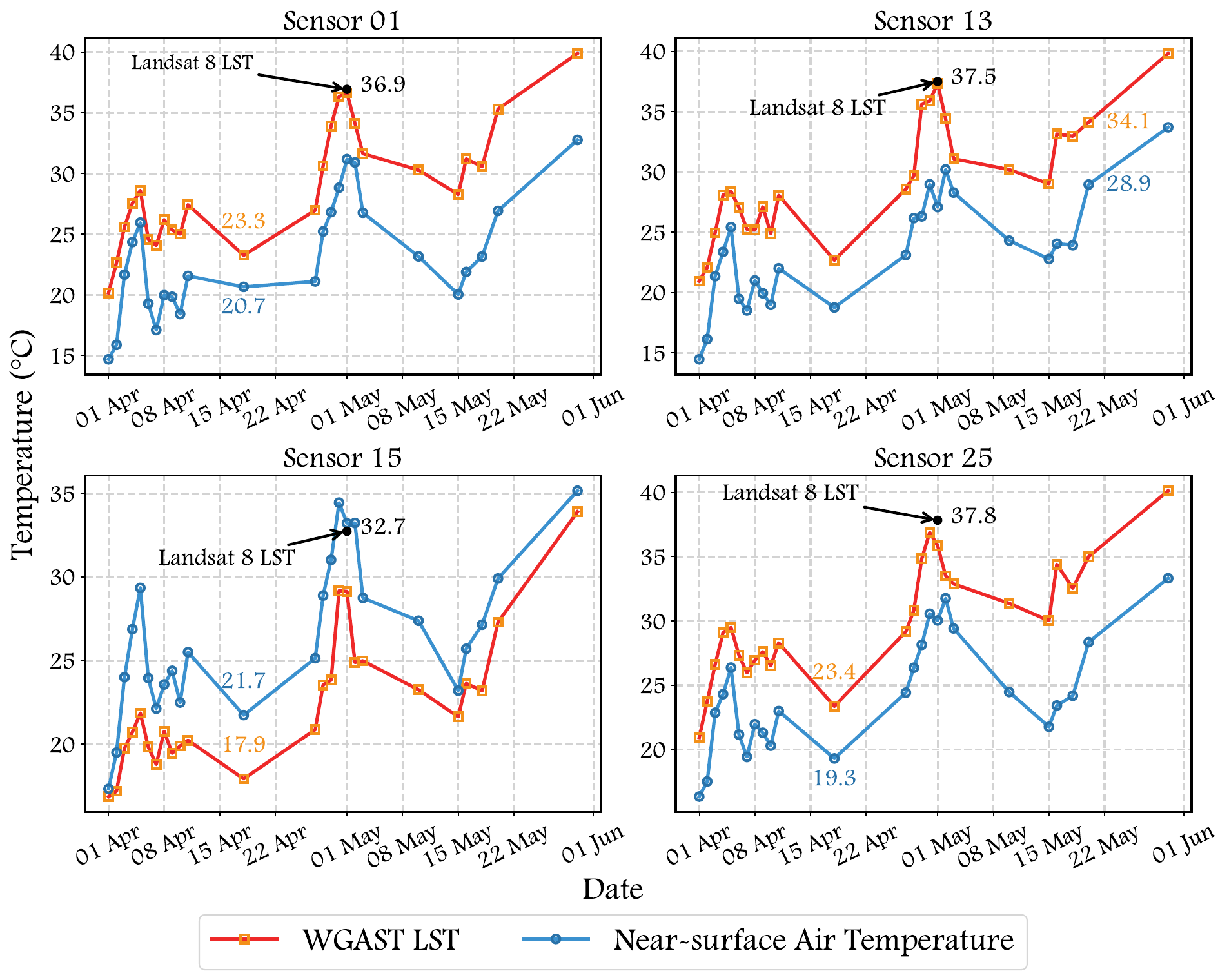}
  \caption{Scatter plots comparing WGAST’s 10 m LST with near-surface air temperature \(T_a\) from four randomly selected sensors during the evaluation period. The plots illustrate the strong correlation and synchronized temporal patterns between the generated LST and \(T_a\).}
 
  \label{fig:scatter_eval}
\end{figure}

Fig.\ref{fig:scatter_eval} displays scatter plots comparing WGAST’s 10 m fused LST with near-surface air temperature (\(T_a\)) from four randomly selected sensors during the evaluation period.  These plots demonstrate a strong correspondence between LST and \(T_a\), confirming that the fused LST product reliably captures the physical trends observed at ground level. The temporal evolution of LST and \(T_a\) shows a clear and synchronized pattern across all sensors. In nearly all cases, both LST and \(T_a\) exhibit the same monotonic behavior, with LST rising as \(T_a\) increases and following the same downward trend as \(T_a\) decreases. Even during periods of sharp thermal fluctuation, WGAST's LST closely tracks \(T_a\) by accurately detecting peaks and valleys across diverse spatial settings. Since the selected sensors span various land cover types, these results underscore the model’s robustness and generalizability across different environments. In addition, we included the Landsat 8 LST acquisition on 01 May 2025, which was the only image partially usable with ~20\% cloud-free coverage. All other Landsat acquisitions during the evaluation period were either fully cloud-covered or affected by technical issues. The Landsat 8 LST values are represented as black circles in the plots. While a single Landsat observation is insufficient for full quantitative validation, it provides an independent qualitative reference, that demonstrates that the fused 10 m LST values from WGAST are broadly consistent with available satellite measurements. For instance, for sensor 15, when WGAST’s LST is below $T_a$, it was also the case for Landsat 8, and vice versa. Most importantly, WGAST achieves continuous LST estimates even on dates when Landsat data are missing. Overall, the analysis validates the WGAST’s ability to produce temporally coherent and physically realistic LST estimates in strong agreement with ground truth observations.

\begin{figure*}[h]
  \centering
  \includegraphics[width=0.88\textwidth]
  {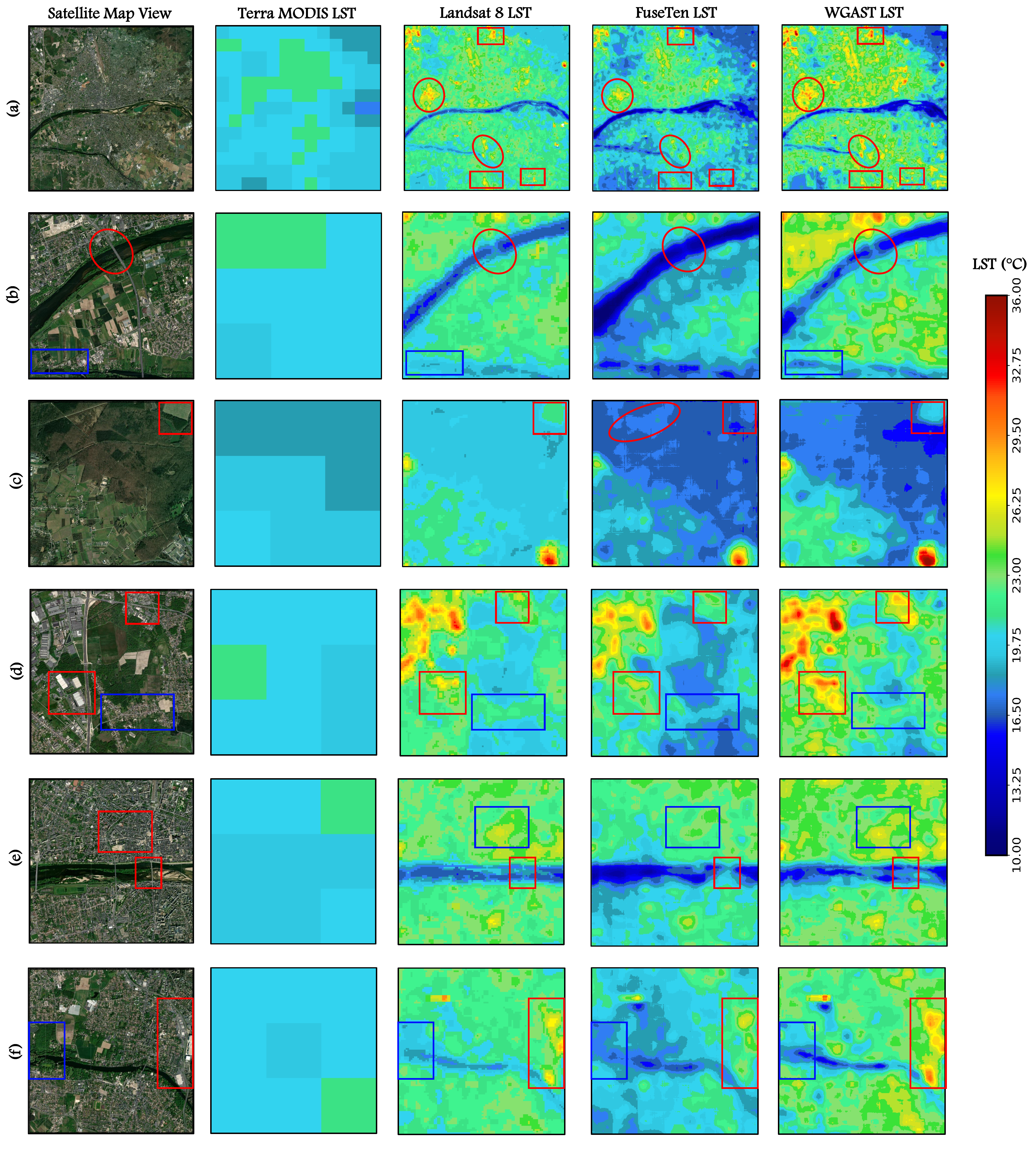}

  \caption{Qualitative comparison of WGAST and FuseTen across six representative regions in Orléans Métropole on 21 Oct 2024: (a) the Orléans Métropole, (b) a semi-urban corridor along the Loire River, (c) the Orléans Forest, (d) a major industrial area, (e) the city center of Orléans, and (f) a mixed residential and vegetated neighborhood traversed by the Loiret River. Each row displays the high-resolution satellite view alongside the Terra MODIS LST, the Landsat 8 LST reference, and the predictions from FuseTen and WGAST.}

  \label{fig:qualitative}
\end{figure*}

\begin{figure}[h]
  \centering
  \includegraphics[width=0.48\textwidth]{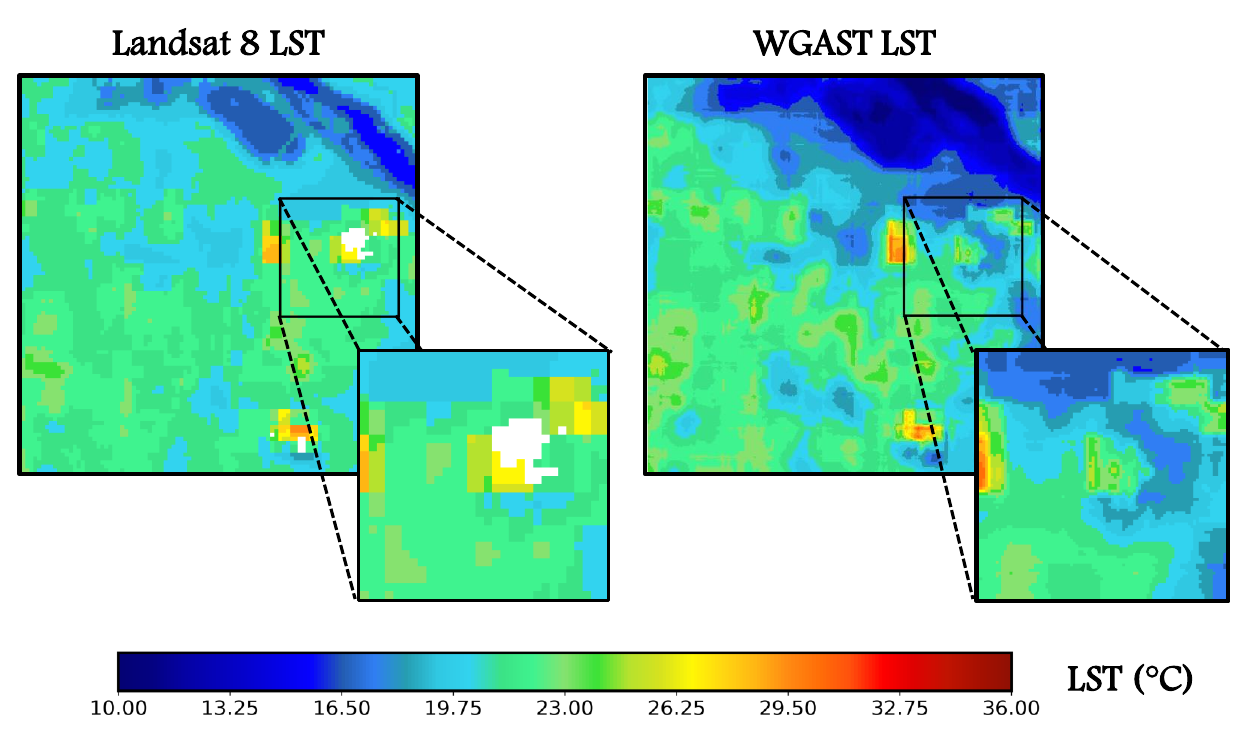}
  \caption{WGAST reconstruction of cloud-covered area in Landsat 8 LST on 21 Oct 2024. The comparison shows Landsat 8 LST (left column) and the corresponding WGAST LST prediction (right column). Zoomed-in areas highlight regions with missing data in Landsat due to clouds.}
  \label{fig:qualitative2}
\end{figure}

\vspace{-1em}
\subsection{Qualitative Assessment}

Fig.\ref{fig:qualitative} presents a qualitative comparison on 21 Oct 2024 between the 1 km Terra MODIS LST, the reference 30 m Landsat 8 LST, the 10 m LSTs generated by FuseTen and WGAST, and the corresponding high-resolution satellite image for context. Other baselines are omitted due to their inferior performance (Table~\ref{tab:average}).

In Fig.\ref{fig:qualitative}(a), WGAST produces LST distributions that closely match the Landsat 8 LST reference while preserving spatial coherence and fine thermal details. Around urban hotspots (highlighted in red), WGAST accurately captures temperature gradients and sharp river boundaries, whereas FuseTen oversmooths these transitions, losing key spatial and thermal contrasts. In semi-urban areas along the Loire and Loiret rivers, WGAST successfully reconstructs cool water surfaces and surrounding structures, including bridges and secondary channels, which are poorly defined in FuseTen. In Fig.\ref{fig:qualitative}(b), WGAST recovers the Loire River’s cool thermal signature, its curved geometry, and surrounding landscape features. The bridge crossing the river (highlighted in red) is sharply delineated in WGAST, appearing more distinct than in both the Landsat 8 LST, where it is faint, and FuseTen, where it is nearly invisible due to oversmoothing. WGAST also captures the secondary Loiret River (highlighted in blue) with well-defined riverbanks and cooler temperatures relative to nearby urban areas. This thermal and structural contrast is hardly visible in the Landsat 8 LST. Fig.\ref{fig:qualitative}(c) focuses on a densely forested region of the Orléans Forest, the largest forest in France. WGAST delivers a detailed reconstruction of the forest thermal structure by accurately capturing subtle intra-canopy temperature variations and preserving crisp spatial textures across the scene. The dense vegetation yields generally lower LST, a pattern faithfully reflected by WGAST. Notably, a small vegetated patch with slightly elevated LST (highlighted in red) is correctly reproduced by WGAST in agreement with the Landsat 8 reference, whereas FuseTen assigns this patch a temperature indistinguishable from the surrounding forest and thus overlooks the anomaly. Fig.\ref{fig:qualitative}(d) depicts a highly active industrial zone characterized by industrial facilities, dense road networks, and fragmented green spaces. WGAST accurately recovers thermal hotspots (highlighted in red) that align closely with the Landsat 8 reference by preserving the contrast between built-up and vegetated areas. In particular, a cluster of buildings showing elevated temperatures (highlighted in blue) is successfully reconstructed by WGAST with markedly higher LST than adjacent vegetated surfaces. By contrast, FuseTen excessively smooths the temperature field, merging distinct thermal patterns and obscuring land-use contrasts. It also fails to emphasize the hotspots and instead produces nearly uniform temperatures across industrial and non-industrial zones. Fig.\ref{fig:qualitative}(e) presents the city center of Orléans Métropole, a densely built urban area traversed by the Loire River, with numerous roads and bridge-like structures. WGAST preserves urban morphology and accurately captures elevated LST in urban cores, particularly in the region highlighted in blue, consistent with the Landsat 8 reference. Linear features such as roads and bridges (highlighted in red) are sharply delineated in WGAST while maintaining thermal contrast with surrounding surfaces. The Loire River itself remains clearly defined spatially and thermally, whereas FuseTen underestimates urban LST, blurs structural details, and struggles to reproduce the river and linear features. Fig.\ref{fig:qualitative}(f) depicts a heterogeneous urban landscape of small residential areas, streets, and vegetated spaces, intersected by the Loiret River. WGAST accurately reflects spatial diversity, separating cooler vegetated zones from warmer surfaces. In the red highlighted area, a localized hotspot is correctly reconstructed, whereas FuseTen oversmooths it. In the blue highlighted region corresponding to the river and its banks, WGAST preserves the expected thermal gradient, clearly distinguishing cooler water from warmer land, while FuseTen assigns nearly uniform temperatures that obscure this contrast.

Overall, WGAST produces more physically coherent and realistic LST than FuseTen. It preserves fine spatial structures and thermal gradients, but also generates detailed daily 10 m LST that even surpasses the 30 m Landsat reference, all derived from the coarser 1 km Terra MODIS input.

Moreover, WGAST effectively mitigates the cloud gaps commonly present in Landsat 8 LST products, as illustrated in Fig.\ref{fig:qualitative2}. This is achieved by relying solely on Terra MODIS LST at the target time, which is less affected by persistent cloud cover. Consequently, WGAST generates 10 m LST that are temporally aligned with Terra MODIS acquisitions and largely free from the missing data affecting Landsat 8. Note that the Terra MODIS input should itself be cloud-free to achieve these results.

\subsection{Spatio-Temporal Generalization}

To evaluate the spatio-temporal generalization capability of WGAST, we tested it on six additional regions with diverse geographic and climatic characteristics: Tours and Montpellier (France), Madrid (Spain), Rome (Italy), Cairo (Egypt), and Istanbul (Turkey). These regions span climates from temperate oceanic (Tours), Mediterranean (Montpellier, Rome), continental Mediterranean (Madrid), arid desert (Cairo), to transitional Mediterranean-continental (Istanbul). WGAST predictions were generated for 21 Oct 2024 (Tours), 01 Apr 2025 (Montpellier), 28 Mar 2025 (Madrid), 10 Jun 2025 (Rome), 23 Dec 2023 (Cairo), and 10 Apr 2025 (Istanbul), representing seasons and thermal conditions unseen during training. The ROIs used for these tests were defined as follows. Tours extends from 47$^\circ$20'51.49''N to 47$^\circ$27'08.74''N and from 0$^\circ$36'57.71''E to 0$^\circ$46'13.20''E. Montpellier spans 43$^\circ$33'27.54''N to 43$^\circ$39'52.00''N and 3$^\circ$48'09.59''E to 3$^\circ$56'58.73''E. Rome covers 41$^\circ$50'11.91''N to 41$^\circ$56'29.58''N and 12$^\circ$25'20.64''E to 12$^\circ$33'46.36''E. Cairo extends from 29$^\circ$59'17.37''N to 30$^\circ$05'40.83''N and 31$^\circ$09'54.75''E to 31$^\circ$17'15.55''E. Madrid spans 40$^\circ$21'42.09''N to 40$^\circ$28'07.75''N and 3$^\circ$46'34.55''W to 3$^\circ$38'09.95''W. Istanbul extends from 41$^\circ$00'05.27''N to 41$^\circ$06'25.35''N and 28$^\circ$57'19.38''E to 29$^\circ$05'41.14''E.

\begin{figure}[h]
  \centering
  \includegraphics[width=0.48\textwidth]
  {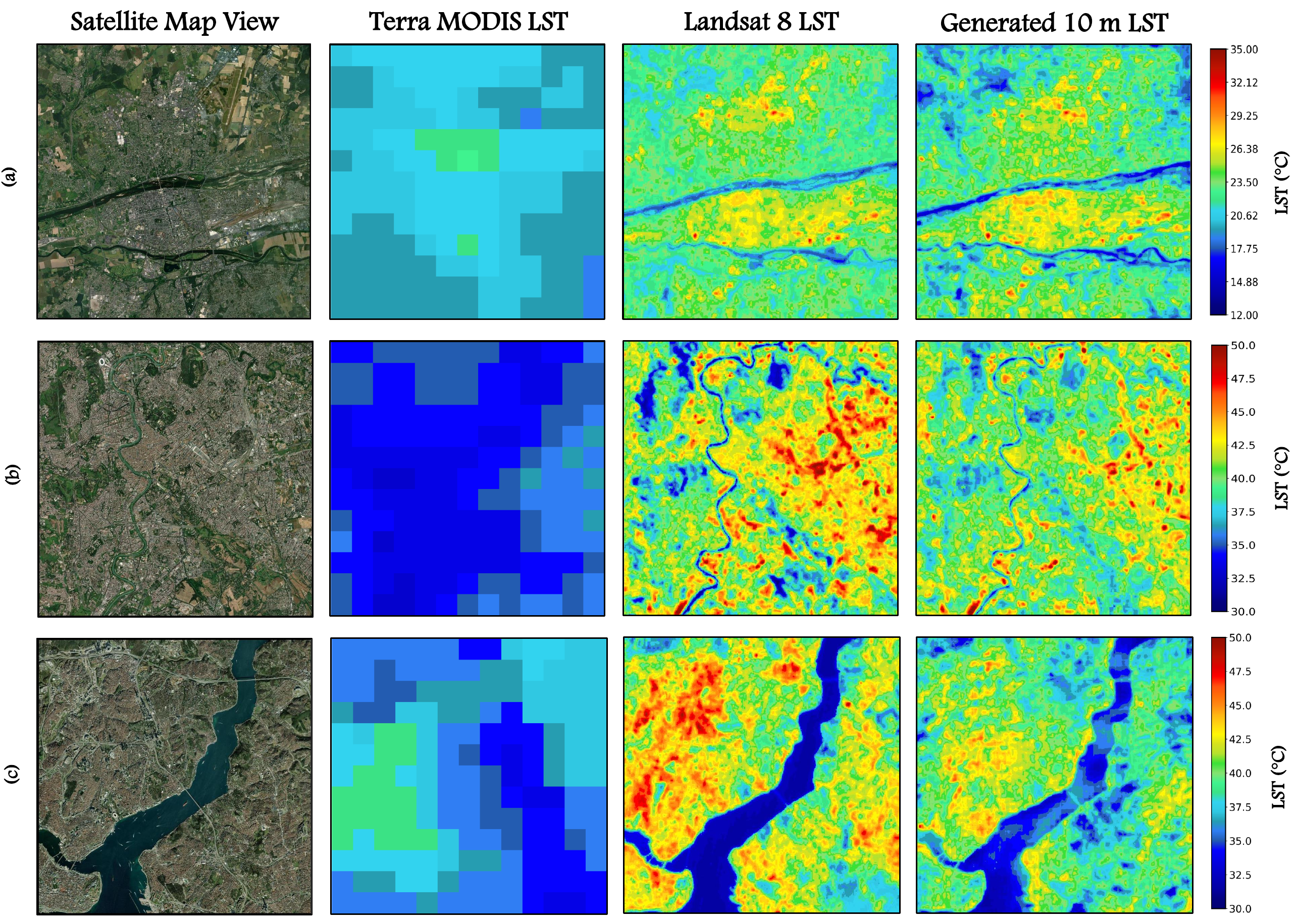}

    \caption{Visual illustration of WGAST predictions across regions with different climatic characteristics and acquisition dates: 
(a) Tours (temperate oceanic, 21 Oct 2024), (b) Rome (Mediterranean, 10 Jun 2025), and (c) Istanbul (transitional Mediterranean-continental, 10 Apr 2025).}

  \label{fig:generalization_visual}
\end{figure}

Table~\ref{tab:new_regions} reports the quantitative performance. Tours, climatically similar to Orléans, shows a low RMSE of 1.099 and high SSIM of 0.858. Madrid maintains comparable performance, while Montpellier and Rome exhibit modestly higher RMSEs (1.743 and 1.786, respectively) but remain reliable. In regions with climates markedly different from Orléans, Cairo and Istanbul, RMSE increases (2.689 and 2.767) yet remains moderate and acceptable, especially given that baseline methods recorded higher RMSEs even within the training ROI (Orléans Metropole).  Overall, these results confirm WGAST’s strong spatio-temporal generalization across diverse climates and geographies by effectively capturing thermal patterns and fine spatial structures, such as the Loire River in Tours, the Tiber River in Rome, and the Bosphorus Strait in Istanbul, as illustrated in Fig.\ref{fig:generalization_visual}.
\vspace{-0.5em}

\setlength{\tabcolsep}{3pt} 
\begin{table}[h]
\centering
\caption{Model performance in new regions.}
\renewcommand{\arraystretch}{1} 
\small

\begin{tabular}{lcccccc}
\toprule
\textbf{Region} & \textbf{RMSE} & \textbf{SSIM} & \textbf{PSNR} & \textbf{SAM} & \textbf{CC} & \textbf{ERGAS} \\
\midrule
Tours       & 1.099  & 0.858  & 25.080  & 2.637  & 0.899  & 1.578 \\
Montpellier & 1.743  & 0.793  & 21.095  & 2.901  & 0.791  & 2.394 \\
Madrid      & 1.178  & 0.872  & 26.192  & 2.821  & 0.821  & 1.702 \\
Rome        & 1.786  & 0.829  & 22.523  & 2.364  & 0.817  & 1.451 \\
Cairo       & 2.689  & 0.848  & 16.429  & 4.208  & 0.771  & 4.211 \\
Istanbul    & 2.767  & 0.724  & 15.712  & 7.469  & 0.765  & 5.156 \\
\bottomrule
\end{tabular}
\label{tab:new_regions}
\end{table}

\subsection{Ablation Study}
We focus on evaluating the contribution of input modalities to WGAST’s performance. Specifically, we investigate how the inclusion or exclusion of different spectral indices, NDVI, NDWI, and NDBI, affects the model’s ability to predict daily 10 m LST. For clarity, when we refer to a given index, we include both the 10 m Sentinel-2 and 30 m Landsat 8 indices, along with all associated encoding components in the generator. Table~\ref{tab:ablation_study} summarizes the results. It is evident that the full combination of NDVI, NDWI, and NDBI yields the best performance across most quantitative metrics, which demonstrates that each index provides complementary information that enhances spatial fidelity, structural accuracy, and spectral consistency. Removing or partially excluding indices generally leads to higher RMSE and ERGAS and lower SSIM and PSNR.

\setlength{\tabcolsep}{2.5pt} 
\begin{table}[h]
\centering
\caption{Ablation study on input modalities.}
\vspace{-0.5em}
\renewcommand{\arraystretch}{1} 
\small

\begin{tabular}{lcccccc}
\toprule
\textbf{Input} & \textbf{RMSE} & \textbf{SSIM} & \textbf{PSNR} & \textbf{SAM} & \textbf{CC} & \textbf{ERGAS} \\
\midrule
NDVI       & 2.726  & 0.849  & 19.468  & 4.226  & 0.765  & 3.657 \\
NDWI       & 2.375  & 0.832  & 19.440  & 3.624  & 0.813  & 3.301 \\
NDBI       & 2.633  & 0.855  & 20.511  & 4.062  & 0.829  & 3.513 \\
NDVI/NDWI  & 2.082  & 0.856  & 22.429  & 4.101 & 0.810  & 2.712 \\
NDVI/NDBI      & 1.960  & 0.851  & 18.903  & 3.998  & 0.832  & 3.724 \\
NDWI/NDBI   & 2.747  & 0.845  & 18.903  & 3.988  & 0.803  & 3.724 \\

\rowcolor{gray!15}
NDVI/NDWI/NDBI   & 1.951  & 0.865  & 23.704  & 4.148 & 0.830 & 2.582 \\
\bottomrule

\end{tabular}
\label{tab:ablation_study}
\end{table}

\section{Conclusion}
In this work, we introduced WGAST, a weakly-supervised generative network for daily 10 m LST estimation via STF of Terra MODIS, Landsat 8, and Sentinel-2 data. To the best of our knowledge, WGAST is the first fully DL–based framework specifically designed for this task. The framework is built upon a cGAN, where the generator follows a four-stage pipeline: feature extraction, feature fusion, LST reconstruction, and noise suppression. A key contribution lies in using cosine similarity to guide fusion by transferring relevance scores from Sentinel-2 and Landsat 8 spectral features to Landsat 8 LST features. Training is weakly supervised through physical averaging, where the generated 10 m LST is upsampled to 30 m and compared to Landsat 8 LST. WGAST requires only Terra MODIS LST at the target time and a prior triplet $T_1$ combining Terra MODIS, Landsat 8, and Sentinel-2, thereby preserving daily temporal resolution. WGAST outperforms existing methods both quantitatively and qualitatively. Compared to FuseTen, it achieves on average a 17.05\% reduction in RMSE, 4.22\% improvement in SSIM, 10.98\% increase in PSNR, and 13.76\% reduction in ERGAS, with even larger gains over traditional linear and existing combinations of STF and downscaling methods. It also preserves fine spatial structures such as temperature gradients along rivers and bridges, variations around buildings, and detailed urban and agricultural patterns. Additionally, by relying solely on Terra MODIS LST at the target date, WGAST effectively overcomes cloud-induced gaps in Landsat 8, by producing complete, high-resolution, and physically consistent LST maps. Its robustness is further confirmed by the strong agreement between the generated 10 m LST maps and in situ measurements from 33 near-ground sensors across the ROI.

WGAST demonstrates strong spatio-temporal generalization across diverse regions and climates. However, in areas with climatic conditions that differ substantially from the training region, a noticeable yet moderate performance drop occurs. Future work will focus on developing an inference and transfer learning framework to enable automatic adaptation to unseen regions without requiring full retraining. Furthermore, regarding practical implementation, generating daily 10 m LST with WGAST requires two inputs: Terra MODIS LST at the target date and a reference triplet comprising Terra MODIS, Landsat 8, and Sentinel-2 observations at a nearby date with minimal cloud coverage. Missing pixels are filled using adaptive spatial interpolation. Future direction will explore extending WGAST to explicitly handle incomplete inputs by enabling the model to predict and fill gaps directly during inference.



\vspace{-1em}

\vspace{1em}
\bibliographystyle{IEEEtran}
\bibliography{references}

\begin{IEEEbiography}[{\includegraphics[width=1in,height=1.25in,clip,keepaspectratio]{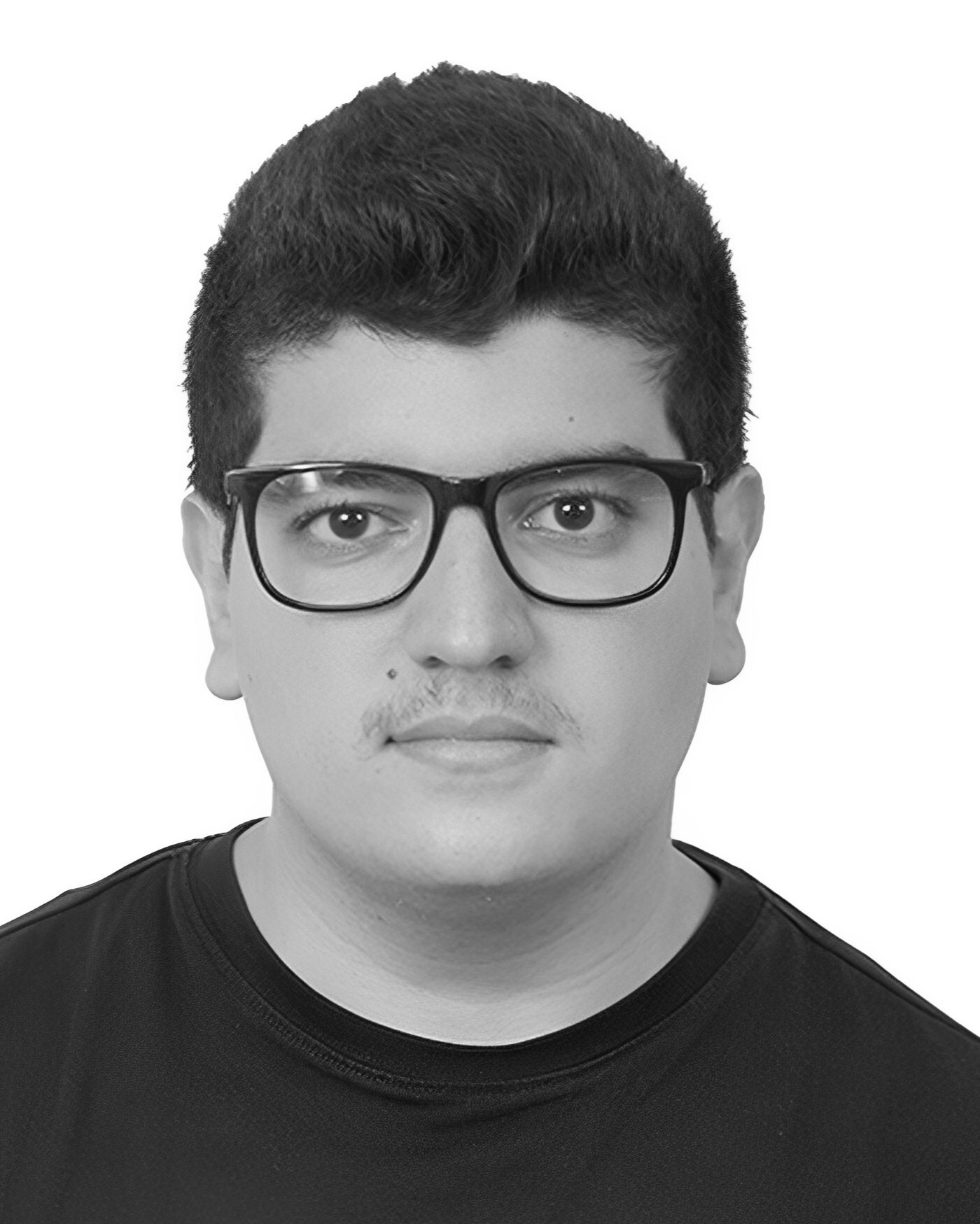}}]{Sofiane Bouaziz}

is PhD Student at INSA Centre Val de Loire (INSA CVL) and the PRISME Laboratory. He received his Master's and Engineering degrees in Computer Science and Artificial Intelligence from École Nationale Supérieure d'Informatique, Algiers, Algeria in 2023. His current research focuses on using Artificial Intelligence and Computer Vision to tackle challenges in remote sensing.
\end{IEEEbiography}

\vspace{-8em}
\begin{IEEEbiography}[{\includegraphics[width=1in,height=1.25in,clip,keepaspectratio]{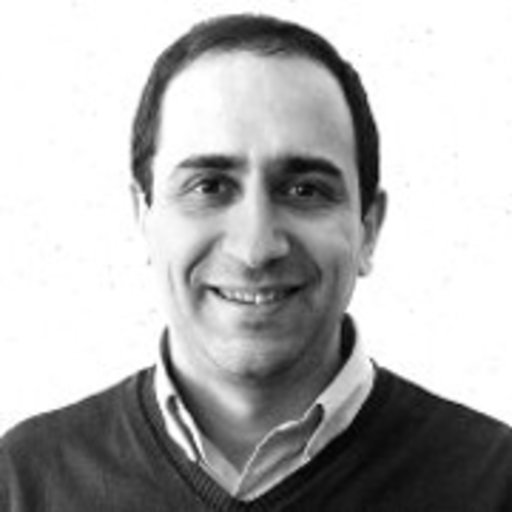}}]{Adel Hafiane}
received the M.S. degree in embedded systems and information processing, and the Ph.D. degree from the University of Paris-Saclay, in 2002 and 2005, respectively. He subsequently spent a year engaged in teaching and research. This followed by postdoctoral position at the computer science department, University of Missouri, from 2007 to 2008. Since September 2008, he joined INSA CVL as associate professor and later the university of Orléans in 2025. He is a head of the Image and Vision Group at the PRISME Laboratory, a joint research unit of the University of Orléans and INSA CVL. He was an invited researcher at the University of Missouri on multiple periods, from 2009 to 2013. His research interests include theory and methods of machine learning and computer vision for different applications. He coordinated several research projects and co-authored more than 90 papers and 3 patents. He has also served as an associate editor for several special issues.

\end{IEEEbiography}

\vspace{-10em}
\begin{IEEEbiography}[{\includegraphics[width=1in,height=1.25in,clip,keepaspectratio]{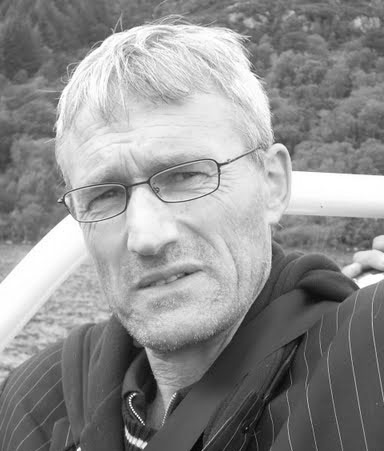}}]{Raphaël Canals}

received the Dipl.Ing. degree in electrical engineering and the Ph.D. degree in electronics from the University of ClermontFerrand, France, in 1989 and 1993, respectively. In 1993, he was Postdoctoral Fellow at the Computer Science Department, CNRC, Ottawa, ON, Canada. In 1994, he joined the Polytechnic School, University of Orléans, France, as a Teacher. He is currently a Researcher with Laboratory PRISME, University of Orléans-INSA CVL. He is also an Associate Professor with the University of Orléans. In 2015, he was introduced at the AgreenTech Valley Cluster dedicated to digital technologies for plant industry. His current interests are in biomedical imaging, innovation for agriculture, the IoT, and AI.

\end{IEEEbiography}

\vspace{-10em}
\begin{IEEEbiography}[{\includegraphics[width=1in,height=1.25in,clip,keepaspectratio]{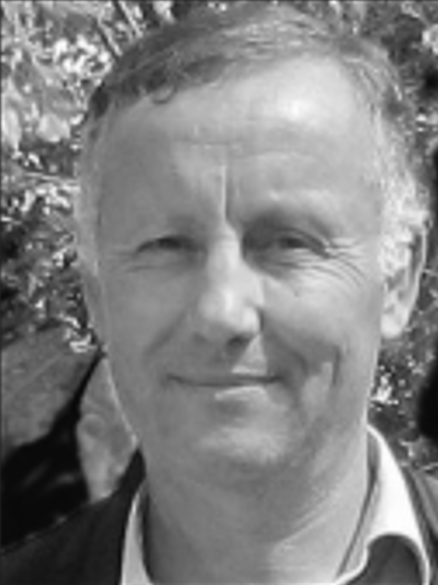}}]{Rachid Nedjai}

is a professor in limnology and geomatics at the University of Orléans, where he also leads the Master's program in Geographic Information Systems (GAED). He obtained his Ph.D. in geochemistry of lacustrine waters and paleoenvironmental reconstruction from the University of Grenoble 1. His research focuses on environmental data simulation, hydrogeology, and geomatics applications for water resource management. He has contributed to numerous international projects, including the redesign of Algeria's judicial map and a spatial data infrastructure for water management in Africa. He has supervised multiple theses on hydrology, water resource management, and environmental risk assessment, with many of his students successfully entering the professional world. His expertise in geomatics and water resource management has made him a key contributor to advancing sustainable practices in the field.
\end{IEEEbiography}

\end{document}